# ABSTRACT

PAHWA, DEEPAK. Decentralized Decision Making in Two Sided Manufacturing-as-a-Service Marketplaces. (Under the direction of Dr. Binil Starly).

Advancements in digitization have enabled two sided manufacturing-as-a-service (MaaS) marketplaces which has significantly reduced product development time for designers. These platforms provide designers with access to manufacturing resources through a network of suppliers and have instant order placement capabilities. Suppliers get access to designers nationwide and compete in a much larger market compared to their traditional markets. Two key decision making levers are typically used to optimize the operations of these marketplaces: pricing (for incoming orders) and matching (allocation of orders to suppliers). The existing marketplaces operate in a centralized structure where they have complete control over decision making. However, a decentralized organization of the platform enables transparency of information across clients and suppliers. This dissertation focuses on developing tools for decision making enabling decentralization in MaaS marketplaces.

In pricing mechanisms, a data driven method is introduced which enables small service providers to price services based on specific attributes of the services offered. A data mining method recommends a network based price to a supplier based on its attributes and the attributes of other suppliers on the platform. Three different approaches are considered for matching mechanisms. First, a reverse auction mechanism is introduced where designers bid for manufacturing services and the mechanism chooses a supplier which can match the bid requirements and stated price. The mechanism ensures that suppliers do not directly compete for price and every supplier gets an equal opportunity to respond to the bid. Since these marketplaces consist of a network of independent suppliers which act rationally and to maximize their payoffs, the platforms often struggle to design a system which can achieve stability in matching i.e. no two participants will prefer each other over their assigned match. The second approach uses mechanism design and mathematical programming to develop a stable matching mechanism for matching orders to suppliers based on their preferences. Empirical simulations are used to test the mechanisms in a simulated 3D printing marketplace and to evaluate the impact of stability on its performance. The third approach considers the matching problem in a dynamic and stochastic environment where demand (orders) and supply (supplier capacities) arrive over time and matching is performed online. This problem is solved using a reinforcement learning framework where neural networks are used for function approximation to handle computational intractability introduced by a large state space. The methods developed in this dissertation lay the foundation of a variety of frameworks for decentralized decision making in MaaS marketplaces.

Decentralized Decision Making in Two Sided Manufacturing-as-a-Service Marketplaces

by
Deepak Pahwa

A dissertation submitted to the Graduate Faculty of
North Carolina State University
in partial fulfillment of the
requirements for the degree of
Doctor of Philosophy

Industrial Engineering

Raleigh, North Carolina

2020

APPROVED BY:

_______________________________    _______________________________
Dr. Binil Starly                                            Dr. Yuan-Shin Lee
Chair of Advisory Committee

_______________________________    _______________________________
Dr. Paul Cohen                                            Dr. Jonathan Duggins

# DEDICATION

*To my family and my advisor.*

# BIOGRAPHY

Deepak Pahwa hails from Karnal, Haryana, India. He completed his Bachelor of Technology degree in Mechanical Engineering from National Institute of Technology Kurukshetra in 2010. He worked with Honda Motors and Evalueserve until 2015 when he joined the Edward P. Fitts Department of Industrial and Systems Engineering at North Carolina State University to pursue his PhD in Industrial Engineering. He held several research and teaching assistant positions at NC State. His research interests focus on applications of machine learning, mathematical programming and mechanism design for decision making in MaaS marketplaces.

# ACKNOWLEDGEMENTS

First and foremost, I would like to express my gratitude for my advisor Dr. Binil Starly. I have always admired his ability to come up with brilliant ideas, and the patience and flexibility with which he lets his students pursue those ideas. I want to thank him for the trust he placed in me during the last 5 years. Without his guidance and support this work would not have been possible.

I would also like to extend my gratitude for Dr. Umut Dur who introduced me to the field of mechanism design. His critical evaluations and feedback played a key role in this dissertation. I would like to thank Dr. Reha Uzsoy for his guidance and detailed evaluations of my research. I would also like to thank Dr. Paul Cohen for his valuable feedback. Special thanks to the faculty at NC State, notably Dr. Arnab Maity, Dr. Julie Ivy, Dr. John Baugh and Dr. David Dickey for teaching excellent courses which were crucial for the work in this dissertation.

Thanks to DIME lab members Arun, Atin, Ben, Mahmud, Akshay, Nabeel and Aman for intellectually stimulating discussions which made the time at DIME lab fun and enjoyable. Many thanks to my friends Ahmad, Avik, Priya, Amit, Heramb, Kshitij and Kaustubh for the memorable experiences in Raleigh. Additionally, thanks to my former colleague and friend Ravi Singh, who has always been a source of encouragement.

# TABLE OF CONTENTS

# LIST OF FIGURES

# LIST OF TABLES

# Chapter 1 : Introduction

**1.1 Background**

Manufacturing is an essential component of the US economy. In 2018, it contributed 11.6% to the US economic output. As per the Bureau of Labor Statistics, manufacturing has created 12.5 million direct and 17.1 million indirect jobs in the US. Supporting sectors such as retailing and logistics also benefit from manufacturing industry. Each dollar spent in manufacturing leads to 1.89$ worth of business growth in the supporting sectors. In the era of fourth industrial revolution, advancements in internet connectivity, computing and data storage capabilities are driving new business models leading to transformation in manufacturing Industry. Manufacturing data has become a major digital asset of value in the global manufacturing network [1-6]. However, its impact on small and medium scale manufacturers is still minimal. Since, small and medium scale manufacturers contribute more than 70% of all US manufacturing, the fourth industrial revolution can't be successful without their participation. These manufacturers have limited skills and budgets to leverage the digital advancements. Therefore, technologies need to be developed to support them with tools required to derive value from manufacturing data.

**1.2 Manufacturing-as-a-service Marketplaces**

Digitization has led to development of new business models in manufacturing. One of the prominent models is manufacturing-as-a-service (MaaS) marketplace platforms [7-10] which connect design owners who need a part fabricated (designers) with a fabricator (suppliers) connected to the platform's network [11-13]. In the traditional marketplace, a designer goes through the following steps to procure manufacturing services: 1) contact the suppliers pre-approved by the designer's organization to request quotes 2) share the design and other product specifications such as required resolution with the suppliers 3) receive quotes from multiple suppliers, compare prices and negotiate on due dates or prices 4) finalize a supplier to conduct business with and then sign a contract 5) receive the manufactured product. Suppliers in the pre-approved list of designer's organization typically go through an extensive auditing process before being added to the list. The process consists of site visits to verify the equipment, validation of quality/ other certifications of supplier. If the designer does not find a capable supplier in the list of pre-approved suppliers, it needs to find suppliers from sources like Thomasnet where again validating the supplier's capabilities requires additional time. This multi-step process, often called the RFQ (request for quotes) process, of procuring manufacturing services from a limited set of suppliers often takes weeks and results in long product development cycles.

MaaS platforms significantly shorten this process by removing frictions from the procurement process. Designers upload their 3D design, manufacturing requirements such as material required, preferred manufacturing process and part resolution on the online platform, receive instant price and delivery time quote and place the order immediately. Designers also benefit as they access a variety of manufacturing resources with available capacities through the platform at the click of a button. Suppliers register their machines on the marketplace and receive access to the market nationwide in contrast to their traditional marketplace confined to their local geographic region. They also save on marketing budgets and other resources as orders from the marketplace almost eliminate the RFQ process. Figure 1-1 describes the two sided online MaaS marketplace and lists some of the currently existing marketplaces. MaaS marketplaces operate under two different frameworks connecting designers and suppliers:

**Centralized Platforms:** In a completely centralized operation, the key decision making lies with the platform which determines the price quotes and delivery dates for the incoming orders. It also determines the supplier which an incoming order is allocated to and the payoff to the supplier for an order. The designers and suppliers do not directly interact and the platform is responsible and liable for the parts manufactured by the suppliers.

**Decentralized Platforms:** These platforms provide designers with a list of suppliers which can fulfill their requirements. Designers compare processes and material specifications from various service suppliers which require significant time and effort to evaluate price quotations and historical performance records of suppliers. Decentralized organization of the platform enables transparency of information across clients and end-users. Critical decisions such as whether or not to respond to a job order request, and how to quote price and delivery time for order requests, resides solely with the fabricators.

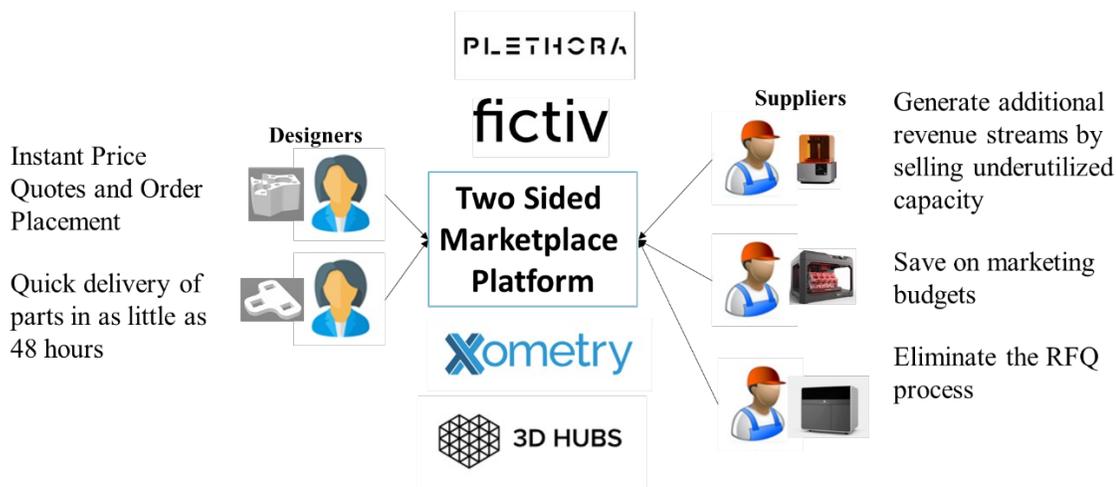

Figure 1-1: Two sided online manufacturing-as-a-service marketplaces

The centralized and decentralized operation of the marketplace explained above can be considered to be two ends of a spectrum. Different frameworks can be adopted to tilt the platform operation towards centralization or decentralization. For instance, a platform operating under a completely centralized operation as described above can take a step towards decentralization by revealing the identities of suppliers to designers. A completely decentralized operation can bring in some level of centralization by shortlisting a potential list of suppliers for an order and recommending it to the designer.

The existing MaaS marketplaces are highly centralized in their operation where they determine the price to charge a designer and the payoff for the supplier as well. The suppliers and the designers do not directly interact and they rely on the platform for any communication. The platform also determines the supplier to which an order would be allocated. The designers and suppliers have very limited role in decision making especially concerning pricing and allocation.

**1.3 Research Motivation**

Though MaaS platforms benefit the suppliers and designers as discussed in the previous section, their sustainability depends on their ability to change as the market for internet intermediaries in manufacturing evolves. The platform's key source of efficiencies, reducing search costs for designers and sales prospecting costs for fabricators, could be threatened if designers or fabricators perceive that they could extract more value from transactions that bypass the intermediary—i.e., cutting out the middleman [8]. In many regulated products, the identity of the fabricator must be known before a contract is awarded. Hence it is critical that a sustainable intermediary platform reveal the identity of both clients and fabricators on either side. In a progressively transparent structure that is more efficiently driven by decentralized information, service prices through the platform would be determined by the interactions of clients and fabricators, not solely by the platform. A more efficient and sustainable model is achieved when critical decisions reside solely with the fabricators. However, the fabricators (typically small/medium job shops) lack the decision tools to accept/reject jobs or price them as they come in, limiting their ability to participate profitably in these marketplaces.

MaaS marketplaces have two key levers which they adjust for operational efficiency and which determine the extent of decentralization in operations. These are pricing i.e. determine the price for an incoming order and matching i.e. allocation of incoming orders to suppliers. The focus of this dissertation is developing decision making tools enabling decentralization in these two areas in MaaS marketplaces:

1. **Pricing**: Centralized platforms utilize advanced heuristics and data mining algorithms to determine a price offering. The platform decides the price for a particular job order and the service provider simply picks job orders to take on, much like ride-sharing platforms. Decentralized platforms, however, let machine asset owners to set prices. Decentralized manufacturing systems are often hard to implement, with a major drawback being that price determination by service providers can be ad-hoc and inefficient. In an online marketplace, suppliers compete in a nationwide supplier network which makes traditional activity based costing methodologies inefficient to strategically price services within the market. Large service providers have trained professionals, competition benchmarks and other tools to optimally determine prices whereas small suppliers lack the skillsets to determine competitive pricing levels. Therefore, there is a need to develop algorithms which can support small and medium scale suppliers to determine the optimal price for their services in order for them to be competitive in online marketplaces.

2. **Matching:** The suppliers connected on MaaS platforms vary widely based on size, years of experience, location and other factors. Typically, suppliers which are large and are located in prominent locations are able to win majority of orders and small suppliers located in remote places are left out. Large suppliers are able to provide lower prices, owing to economies of scale and can potentially win a designer's trust showcasing their size and experience. This can result in the small suppliers leaving the marketplace over time. A few large suppliers serving the majority of orders on the platform eliminates the need of online intermediary and makes it unsustainable in the long run. A key advantage of these platforms is a large pool of suppliers with variety of resources available to designers at the click of a button. Therefore, to be sustainable, the platforms need to ensure that all participants on the platform benefit in order to incentivize them to actively participate in the marketplace. These platforms can only be sustainable when service providers especially small and medium scale providers are able to capture fair share value of their services offered and provided with enough opportunities to be competitive in the marketplace.

**1.4 Dissertation Objectives**

The following describes the four research objectives and briefly describes the approaches taken to achieve those objectives:

**Research Objective 1**: **Develop a price recommendation tool for suppliers in decentralized MaaS marketplaces**

It is explored whether service providers can price their services based on specific attributes of their business and services offered through a data driven approach in a decentralized marketplace. In a decentralized marketplace suppliers maintain an online profile where they market their services and resources. The reputation system of the platform which informs the designers about the historical customer service record of the supplier is also a part of its online profile. Suppliers currently do not systemically link their price to the attributes of their online profile. A data mining method is developed which recommends a price range to a supplier based on its attributes and the attributes of competing suppliers on the platform. A dataset collected from online marketplace '3D Hubs' is used. This objective is addressed in Chapter 2 of this dissertation.

**Research Objective 2**: **Develop a method for designers to name their own price for manufacturing services in a MaaS marketplace**

A reverse auction mechanism is proposed where designers name their own price, i.e. they bid for manufacturing services and the mechanism chooses a supplier which can match the bid requirements and stated price. Designers, especially price conscious designers, benefit by receiving the service at a price less than market value and suppliers obtain access to an alternate sales channel to sell their excess capacity to price conscious designers at a lower price. The mechanism ensures that suppliers do not directly compete for price and every supplier gets an equal opportunity to receive the bid. This objective is addressed in Chapter 3 of this dissertation.

**Research Objective 3**: **Develop a stable matching method for suppliers and designers in MaaS marketplaces based on their preferences**

Chapter 4 focuses on matching mechanisms to allocate orders based on preferences of designers and suppliers enhancing decentralized decision making. Resource allocation in a MaaS marketplace is a bipartite matching problem where the platform performs the task of matching participants on one side with participants on the other. The notion of stability, similar to the stable marriage problem, is considered where no two participants (designers and suppliers) would prefer each other over their matched participant. Mechanism design and mathematical programming is used to develop these methods.

**Research Objective 4**: **Develop a dynamic matching method for suppliers and designers in MaaS marketplaces**

Chapter 5 considers matching in a dynamic and stochastic environment where an agent learns to accept or reject orders for a supplier in a MaaS marketplace considering the currently available orders and capacities and expected arrivals of orders and capacities in the future. The agent uses a deep reinforcement learning method called deep Q-network to learn directly from experience and make order acceptence decisions in MaaS marketplaces.

The work in this dissertation could serve as a building block for future research on data driven pricing and matching in MaaS marketplaces.

# Chapter 2 : Network-based pricing for 3D printing services in two-sided manufacturing-as-a-service marketplace

## 2.1 Introduction

The rapid prototyping and short run production service providers now have access to 'Sharing-Economy' type two-sided platforms, which connect designers requesting services to providers who are able to fulfil order requirements [1]. These platforms have a network of service providers from which they can draw from to satisfy order requirements. Designers in need of 3D Printing services upload designs, receive instant quotes and place orders on these platforms [2]. Several approaches have been adopted by these platforms for providing instant pricing services. One approach is to let machine asset owners to set prices which allows designers to select service providers as per their choice. Other platforms utilize advanced heuristics and data mining algorithms to determine a network based price offering. In the latter case, the platform decides the price for a particular job order. The service provider simply picks job orders to take on, much like ride-sharing platforms.

These platforms centralize the interaction between designers and service providers but are often in control of the prices and the kind of orders the service providers can receive. Ratings, reviews and service type offerings can often lead to a centralized platform driving much of the operations. In decentralized platforms, participants can broadcast services to the entire network of service providers without the need for a 'middle' layer platform. Without a central platform, there is increased transparency, inclusivity and competition for better services.

However, decentralized manufacturing systems are often hard to implement, with a major drawback being that price determination by service providers can be ad-hoc and inefficient.

In an online marketplace, suppliers compete in a national supplier network which makes traditional activity based costing methodologies inefficient to strategically price services within the market. Large service providers have trained professionals, competition benchmarks and other tools to optimally determine prices whereas small suppliers lack the skillsets to determine competitive pricing levels. During the data collection process, price variation among the service providers were significantly high, leading to users having to shop around various service bureaus to get the best price. This extra effort further delays product development times and can discourage designers from prototyping more often.

A naïve approach to competitive pricing for a service provider in an online marketplace would be to compare against features of other suppliers in the neighborhood and then choose a competitive price. Suppliers currently do not systematically link their price to the attributes of their online profile as this manual approach requires significant effort and does not provide precise results. This work seeks to address the question on whether it is possible for small service providers to price services based on specific attributes of their business and services offered through a market driven approach. Publicly available data from the '3D Hubs' marketplace was utilized to build a machine learning model to recommend a price range for services offered by a provider in an online 3D printing marketplace.

**2.2 Literature Review**

Traditional methods to price additive manufacturing services consist of estimating direct and indirect costs such as raw material and energy costs. One of the first cost models proposed by Hopkinson and Dickens [3] considers machine costs, labor costs and material costs. They assume negligible energy costs and consider annual production of the same part on one machine. Ruffo et al. [4] considers a relatively higher impact of the overhead cost of laser sintering technology. Rickenbacher et al. [5] determines cost per part when parts are printed simultaneously using Selective Laser Melting and Lindemann et al. [6] considers the lifecycle cost, including the cost of pre and post processes, of an additive manufactured part in its model. Schröder et al. [7] proposes an activity-based analysis of the process cost and a sensitivity analysis to determine the key cost influencing factors. In a manufacturing marketplace, the pricing should be value driven instead of being cost driven as a large number of suppliers with widely varying cost components participate in the marketplace.

Sharing economy platforms in other industries have price recommendation tools to support the participants on the platform. Airbnb has developed a data mining tool [8] which uses features of a listing (among other parameters such as future demand) such as location, amenities, guest reviews to predict a price range of a listing and associated probabilities of success for a sale. Tang and Sangani [9] predict price category and neighborhood category for Airbnb listings in San Francisco. They use data from the 'Inside Airbnb' project (Inside Airbnb Dataset [10]) to create a Machine Learning classifier to make the predictions. Chen and Xie [11] and Gibbs et al. [12] study the effect of attributes of a listing such as functionality and reputation on price using a hedonic pricing model for listings from Texas and Canada respectively.

In manufacturing, data-driven approaches for cost estimation of 3D models has been carried out by Chan et. al [13]. Druan et al. [14] developed artificial neural network based models to estimate the cost of piping elements during early design phase. The data driven methods proposed by these papers do not consider the impact of competition or the complexities associated with a networked service platform. Sharing Economy platforms in additive manufacturing industry can benefit by adopting data driven approaches to support suppliers with price prediction. To the best of authors' knowledge, this work is the first implementation of a machine learning method to price 3D Printing services for an online manufacturing marketplace.

## 2.3 Methodology

### 2.3.1 Data Collection

Software was built to scrape data from the public profiles of service bureaus on the '3D Hubs' marketplace platform. As of March 2018, there were 29,554 suppliers listed on the marketplace across the globe. Every supplier has a supplier profile which provides information about its service listings and its reputation on the marketplace. A service listing is a unique combination of a 3D printer, material, resolution and a corresponding price. Table 2-1 provides 21 supplier attributes, information for which was gathered from their profiles published on the platform. After cleaning the data, 546 suppliers with 5,469 service listings located in the United States and 1,043 suppliers with 9,808 service listings located in Europe remained relevant for the study.

### 2.3.2 Feature Extraction for Dataset Preparation:

Material, 3D Printer Model, Supplier Location and Supplier Description attributes were not usable in their existing form and had to be quantified/categorized in order to be used in the model as predictor variables.

*Material*: To consider the material as a predictor variable in the dataset, 346 unique materials listed by these suppliers were categorized in 14 categories. Materials were categorized by its type which included: Acrylonitrile Butadiene Styrene (ABS), Polylactic Acid (PLA), Specialty ABS,

Specialty PLA, Polyethylene Terephthalate (PET), Specialty PET, Polycarbonate (PC), Specialty PC, Nylon, Specialty Nylon, Flexible Material (Thermoplastic Elastomer/ Polyurethane), Acrylonitrile Styrene Acrylate (ASA), Metals, Resins, Soluble Material (High Impact Polystyrene, Polyvinyl Alcohol) and Others.

Table 2-1: Listing attributes and their description

| Feature Category | Supplier Feature | Description |
|---|---|---|
| Customer Feedback Features | Average Rating | Service evaluation ratings from customer reviews (1 to 5 with 5 being the best and 1 being the worst). Average Rating is the mean of Print Quality Rating, Speed Rating, Service Rating and Communication Rating |
| | Print Quality Rating | |
| | Speed Rating | |
| | Service Rating | |
| | Communication Rating | |
| | Number of Reviews | Number of customer reviews |
| | Average Response Time | Average time taken to respond to a customer's request |
| Supplier Experience and Scale | Activation Date | Date of supplier's activation on '3D Hubs' |
| | Number of Machines | Number of machines registered on '3D Hubs' |
| | Registered Business | Whether the supplier has a registered business |
| Location | Supplier Location | Country, State, City and GPS coordinates |
| Supplier Description Features | Supplier Description | Business Description of supplier |
| | Print Sample Images | Number of print sample/ other images uploaded on the profile |
| Printer and Material Features | 3D Printer Model | Model of the listed 3D Printer; 3D Printer Cost and 3D Printing Process were derived from the model |
| | Material | 3D Printing Material |
| | Resolution | Print Resolution |
| | Order Completion Time | Number of days required to complete an order |
| Target Feature | Price | Price for a 10 cm tall 3D Printed Marvin Model |

*3D Printer Model*: To consider the impact of the type of 3D printer on the service price, two parameters were considered: cost of the listed 3D printer and type of 3D printing process. The 3D printing process categories were as follows: Fused Deposition Modelling (FDM), Stereolithography (SLA), Laser Sintering (Selective Laser Sintering for polymers and Direct Metal Laser Sintering for Metals) and Jetting (Material and Binder Jetting). 528 unique 3D printers were categorized, and their cost was gathered from online sources.

*Location:* To understand the impact of a supplier's location on its price, GPS coordinates of suppliers were used to create geographic clusters. K-means Clustering was used to generate six supplier clusters in United States and nine supplier clusters in Europe. This algorithm assigns each data point to one of K clusters based on the features (GPS coordinates) that are provided. It assigns the data points to clusters such that the total intra-cluster variation is minimized.

*Supplier Description:* To understand the impact of supplier description on its price and use it as a predictor variable, supplier description feature vectors were created. 100 Suppliers were randomly selected from the dataset and their text description was used to create a dictionary of keywords. The keywords were segmented in to five categories: Design Services (scanning, modelling), Logistics (turnaround time, pick-up, free shipping), Specialties (such as jeweler, dental, medical), Experience (years of experience, profession, education) and Additional services (services such as finishing, polishing, laser cutting). Supplier description feature vectors include count of number of words belonging to each of the five keyword categories. Suppliers in Europe provided their description in regional languages which were translated to English to create the feature vectors.

### 2.3.3 Dataset Statistics

***Correlation***: Correlation between numeric listing attributes (Average Rating, Print Quality Rating, Speed Rating, Service Rating, Communication Rating, Number of Reviews, Average Response Time, Activation Date (number of days since activation), Order Completion Time, Resolution, Number of Machines, Number of Print Sample Images, 3D Printer Cost and Price) from Table 2-1 was measured to determine linear or non-linear monotonic relationship between them. All numeric predictor variables had poor correlation (< 40%) with the target variable Price. Correlation between the predictor variables was also determined to remove highly correlated variables from the set of predictor variables. Consumer ratings (Average, Print Quality, Speed, Service and Communication Rating) had significant correlation (> 90%) between them. However, all other numeric predictor variables had poor correlation (< 40%) between them. Considering the high correlation between the consumer ratings, Print Quality, Speed, Service and Communication Rating, these attributes were dropped from the set of predictor variables. Average Rating and all other predictor variables defined in Table 2-1 were used in the model.

***Distributions:*** Price distribution for service listings was right skewed in both US and Europe with prices varying from $2.36 – $1956 in the US and $3.75 - $2261.5 in Europe. Majority of the listings (84% in US and 89% in Europe) were FDM printers, significantly fewer (13% in US and 10% in Europe) were SLA and rest SLS and Jetting. In terms of material categories, the majority of the listings (54% in US and 57% in Europe) were ABS and PLA materials including specialty formulations. Resins constituted 15% of the listings in US and 10% in Europe, and metals constituted only 0.2% of the listings in both US and Europe. Machine cost also yielded a right skewed distribution with cost varying from $175 to ~$1M. The distributions for these listing attributes suggest that majority of the suppliers own low end machines, use low end materials and sell at the lower end of the price spectrum.

2.3.4 **Model Formulation**

The model was formulated with Price as the target variable and rest of the listing attributes as predictor variables. Average Rating, Number of Reviews, Activation Date (number of days since activation), Average Response Time, Order Completion Time, Number of Machines, 3D Printer Cost, Number of Print Sample Images and Resolution were considered as numeric variables and Registered Business, Supplier Location, 3D Printing Process and Material were considered as categorical variables. Figure 2-1 represents the model formulation.

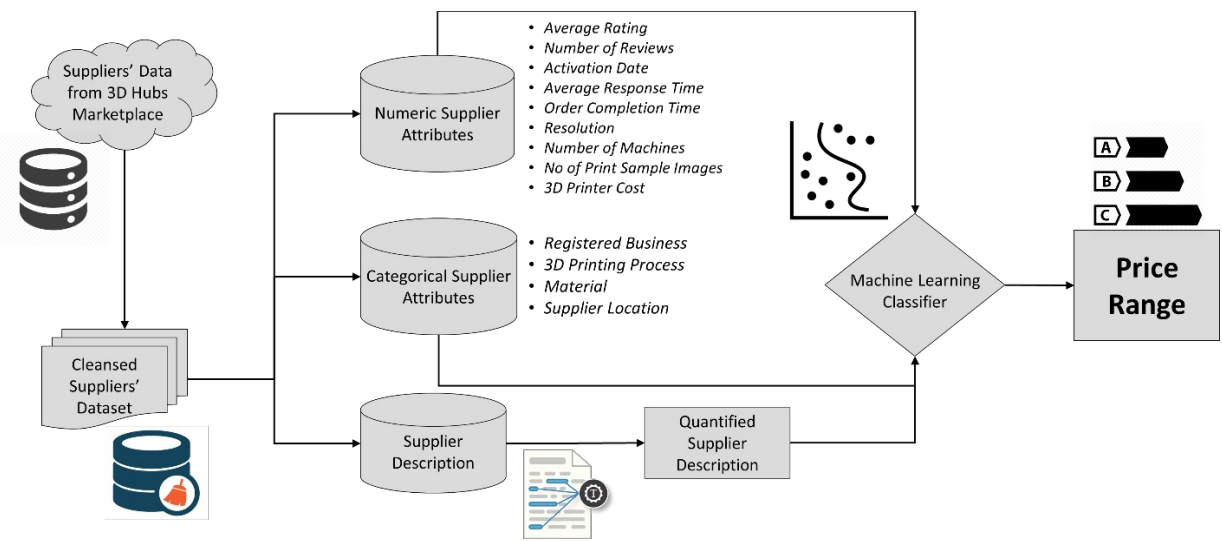

Figure 2-2: Machine Learning Model Formulation

Every supplier on '3D Hubs' listed the price for a 10 cm tall Marvin model for each service listing i.e. unique combination of material, resolution and 3D printer offered by the supplier. Since this was the uniform price metric available for each service listing, it was considered as a price equivalent for service listings' offer. Price was categorized into quartiles resulting in the following four categories in the US: $2.36 - $15.1, $15.1 - $21.2, $21.2 - $36.2, $36.2 - $1956. Since the price distribution was right skewed, the price range was extremely wide in the fourth quartile and it was further divided into quartiles resulting in $36.2 - $47.8, $47.8 - $64.4, $64.4 - $106, $106 – $1956. This led to a total of seven price categories. The last category still had a wide price range, however, due to limited observations in this category, it could not be further divided to narrow down the price range.

The data was divided in to training and test sets in an 80:20 ratio using stratified sampling. After preparing the dataset for a classification model, repeated observations were identified in the dataset. For instance, a repeated observation denotes a supplier having two different service listings which result in same listing attributes after material, 3D printer and price categorization. Repeated observations are also important for the dataset as they represent the real-world environment where a supplier can have two FDM printers with a 200-micron resolution ABS material offering with the same price category. However, they can provide misleading results if the same observation is present in both the test and training sets. To overcome this shortcoming, the observations which were present in both test and training set were removed from the test set. The dataset was fit on a Support Vector Machine (SVM) classifier [15] which solves the following objective function:

$$\min_{w,b,\xi} \frac{1}{2} w^T w + C \sum_{i=1}^{l} \xi_i$$

$$\text{subject to } y_i \left( w^T \phi(x_i) + b \right) \geq 1 - \xi_i$$

$$\xi_i \geq 0, i = 1, 2, \ldots, l$$

$$k(x_i, x_j) = \phi$$

Here $x_i$ represents the vector of listing attributes, $y_i$ represents price classes and $w$ represents the listing attribute weights. $C>0$ is the regularization parameter and determines the sensitivity of the classifier to misclassification. $\phi(x_i)$ maps $x_i$ to a higher dimension and kernel function $k(x_i, x_j)$ determines the hyperplane which separates the different classes.

## 2.4 Results and Discussion

The model was trained on the training dataset using a five-fold cross validation and then tested on the test set. Since the classes were imbalanced, the model adjusts the weight of a class inversely proportional to its frequency. The higher the weight of a class, the higher the penalty for misclassification for that class.

***Grid Search:*** To tune the classifier and select optimal hyper parameters, grid search was performed on Kernel (Radial Basis Function (RBF), Linear, Polynomial and Sigmoid kernels), parameters C and γ. Parameter C behaves as a regularization parameter which trades off misclassification against complexity of the model. Parameter γ defines the region of influence of the support vectors selected by the model. Grid search was performed in two steps. First, a relatively larger range of the parameters ($10^{-4}$ to $10^{4}$) was explored followed by a narrow search around the optimal parameter values found in broader search. RBF Kernel with C = 6500 and γ = 0.01 provided the best results with 72.9% average cross validation accuracy and 65% test set accuracy in the US and 68% average cross validation accuracy and 59.3% test set accuracy in Europe. Grid search provided the optimal hyper parameters to build the classification model.

Table 2-2: Results for SVM Classifier for United States

| Price Range in $ (Price Class) | Train Accuracy | CV Accuracy | Test Accuracy | Precision | Recall | F1 Score |
|---|---|---|---|---|---|---|
| 2.4 - 15.1 (0) | 0.95 | 0.73 (0.01) | 0.80 | 0.71 | 0.8 | 0.75 |
| 15.1 - 21.2 (1) | 0.90 | | 0.62 | 0.62 | 0.62 | 0.62 |
| 21.2 - 36.2 (2) | 0.92 | | 0.61 | 0.67 | 0.61 | 0.64 |
| 36.2 - 47.8 (3) | 0.99 | | 0.43 | 0.44 | 0.43 | 0.43 |
| 47.8 - 64.4 (4) | 0.99 | | 0.48 | 0.48 | 0.48 | 0.48 |
| 64.4 - 106 (5) | 0.99 | | 0.30 | 0.50 | 0.3 | 0.38 |
| 106 - 1956 (6) | 0.98 | | 0.63 | 0.71 | 0.63 | 0.67 |
| Micro Average | 0.93 | | 0.65 | 0.65 | 0.65 | 0.65 |

Table 2-3: Results for SVM Classifier for Europe

| Price Range in $ (Price Class) | Train Accuracy | CV Accuracy | Test Accuracy | Precision | Recall | F1 Score |
|---|---|---|---|---|---|---|
| 3.8 - 15.9 (0) | 0.94 | 0.68 (0.04) | 0.76 | 0.66 | 0.76 | 0.7 |
| 15.9 - 21.5 (1) | 0.87 | | 0.54 | 0.56 | 0.54 | 0.55 |
| 21.5 – 33.1 (2) | 0.91 | | 0.56 | 0.60 | 0.56 | 0.58 |
| 33.1 - 40.1 (3) | 0.99 | | 0.43 | 0.47 | 0.42 | 0.44 |
| 40.1 - 56 (4) | 0.99 | | 0.39 | 0.40 | 0.39 | 0.39 |
| 56 – 87.2 (5) | 0.99 | | 0.45 | 0.57 | 0.45 | 0.5 |
| 87.2 - 2261.5 (6) | 1.00 | | 0.60 | 0.75 | 0.6 | 0.67 |

| Micro Average | 0.93 | | 0.59 | 0.59 | 0.59 | 0.59 |

***Learning Curves:*** Higher values of C can lead to overly complex models resulting in high variance and overfitting. Therefore, to test the model for overfitting, learning curves were plotted. It was found that the model had high variance and was over fitting on the training data with an average training set accuracy of 93% in the US and 92.6% in Europe. To account for overfitting, the model was then tested with lower values of C. C = 100 reduced overfitting significantly with average training accuracy of 78% in the US and 76% in Europe. However, it also reduced both cross validation accuracy and test set accuracy to 65% and 57 % respectively in US and 60% and 53.5% in Europe. Since the model with C = 6500 in US and Europe provided better results on unseen test data, this model was more generalizable and was preferred over models with lower C values.

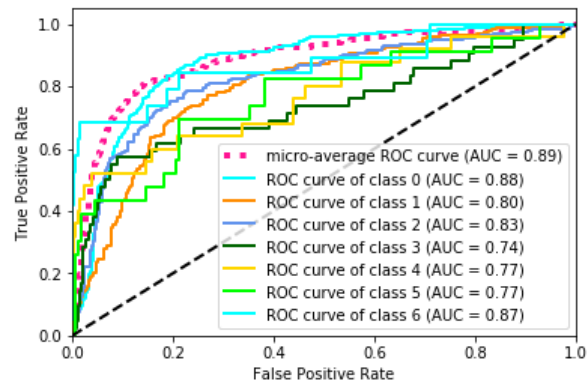

Figure 2-3: ROC Curve for United States (SVM Classifier)

Training accuracies, cross validation accuracies on the training set, test set accuracies, precision, recall and F1 score achieved by the model for each price class are presented in Table 2-2 and Table 2-3. Cross validation accuracies represents the mean and variance (in parenthesis in Table 2-2 and Table 2-3) of accuracies for five folds of cross validation. The baseline accuracy (accuracy with all classes predicted as the class with highest frequency) for this model is 0.25. The micro average test accuracy of 0.65 in US and 0.59 in Europe demonstrates significantly better performance than the baseline.

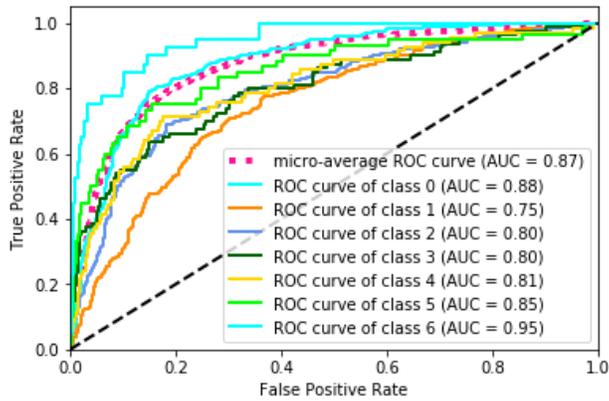

Figure 2-4: ROC curve for Europe (SVM Classifier)

Precision is a measure of exactness of the classifier i.e. the ratio of correctly predicted positive observations to the total predicted positive observations. For example, for price class 0 in the US, 0.71 precision indicates that 71% of the listings which were predicted to belong to class 0, actually belonged to class 0. Recall is a measure of completeness of the classifier i.e. the ratio of correctly predicted positive observations to the all observations in actual class. For example, for price class 0 in the US, 0.8 recall indicates that 80% of the listings belonging to class 0 were correctly predicted to belong to class 0. F1 score provides a balance between precision and recall and is the harmonic mean of precision and recall. Since multi class models have equal number of false positives and false negatives, the micro averages for test accuracy, precision, recall and F1 score are equal. Therefore, these metrics are more important while evaluating performance of each individual class.

The scores for these metrics are higher for the classes 0, 1 and 2 because each of these classes have higher number of observations (25% each) to learn from. Classes 3, 4 and 5 have significantly lower number of observations (6.25% each) to learn from, leading to lower scores for the scoring metrics. Class 6 shows better performance even with lower number of observations because this class has a wide price range with high end machines and materials. This makes class 6 easily separable from rest of the classes.

Receiver Operator Characteristics (ROC) Curves for US and Europe are presented in Figure 2-2 and 2-3. ROC curves illustrate the performance of a classification model at all classification thresholds and 'Area Under the Curve' (AUC) measures the diagnostic ability of the model. The labels were binarized to plot ROC curves for each price class. Since the classes in the models were imbalanced, the micro average ROC curve which provides an aggregate measure of the performance of the model, was also plotted. 0.89 micro average AUC for the US and 0.87 micro average AUC for Europe represents high predictive accuracy of the classifier to differentiate between the price classes.

**2.5 Conclusion**

This work proposes a machine learning approach to determine a network-based price for 3D printing services in an online 3D printing marketplace. The proposed method uses the features of a supplier such as customer reviews and supplier capabilities to predict a price range based on features and prices of other suppliers in the network. The analysis of data from '3D Hubs' marketplace shows that price range of a supplier's listing can be successfully predicted using an array of features extracted from the supplier profiles. The success of data mining-based models for price prediction is only limited by availability of additional data from an online manufacturing marketplace. These marketplaces generate large amount of data which can be used to understand the participants and provide additional value to them. Historical sales data from a manufacturing marketplace could be used to determine the probability of winning an order at a specific price. In addition to supplier's features, order attributes such as 3D design, its due date and designer's attributes such as future potential of orders from the designer can also be considered for price prediction. Impact of additional parameters such as demand forecast, seasonal variation and raw material prices can be added to the model to make it more robust.

# Chapter 3 : Reverse Auction Mechanism Design for the Acquisition of Prototyping Services in a Manufacturing-as-a-Service Marketplace

Pahwa, D., Starly, B., & Cohen, P. (2018). Reverse auction mechanism design for the acquisition of prototyping services in a manufacturing-as-a-service marketplace. Journal of Manufacturing systems, 48, 134-143.

## 3.1 Introduction

The Industrial Internet of Things (IIoT) is ushering in a way of digitization across the manufacturing shop-floor leading to networked production connecting factories and small/medium job shops with clients. IIoT can also enable manufacturing services such as part prototyping to be made more accessible to its users through cloud enabled technologies. Cloud manufacturing technologies can shorten the gap between clients who require prototyping services to service providers who are both capable and available to produce them within the given business and technical constraints. Traditionally, prototyping services bureaus directly offered directly sold their services to the end-user. End-users often had a set of favorite prototyping service bureaus to choose from. Service providers on the other hand spend money and effort marketing their capabilities to both acquire and retain existing customers to help fulfill require number of orders to operate a sustainable business.

Recently, several distributed manufacturing based middleware platforms have emerged to source service providers and deliver prototype parts and short production run parts (CNC prototypes and 3D prints) which match technical constraints against service providers who have the capabilities and capacities to produce them [1-3]. Enabling technology powering such distributed manufacturing platforms include instant pricing algorithms, model based definition (MBD), machine communication protocols and cloud enabled computing software. These platforms make use of unutilized capacity available in the shop-floor of hundreds and thousands of service providers connected to their platform. Two service system organizations have emerged in the matchmaking between clients (those that are require prototyping services) and suppliers (service providers with physical machine assets) have emerged in this prototyping service smart manufacturing marketplace.

***Decentralized Platforms***: These platforms provide end-users with a list of suppliers which can fulfill designer's requirements. Designers compare processes and material specifications from various service suppliers which require significant time and effort to evaluate price quotations and historical performance records of suppliers. Decentralized organization of the platform enables transparency of information across clients and end-users. For instance, 3D Hubs, a 3D printing marketplace platform, has a network of more than 7,000 3D printing suppliers connected globally. The suppliers connected on this platforms vary widely based on size, years of experience, location and other factors. Typically, suppliers which are large and are located in prominent locations are able to win majority of orders and small suppliers located in remote places are left out. Large suppliers are able to provide lower prices, owing to economies of scale and can potentially win a designer's trust showcasing their size and experience. This can result in the marketplace to prefer large service bureaus thus reducing the need of the decentralized platform in the first place. Decentralized platforms can only be sustainable when smaller and distributed service providers are given equal opportunities be competitive in the marketplace.

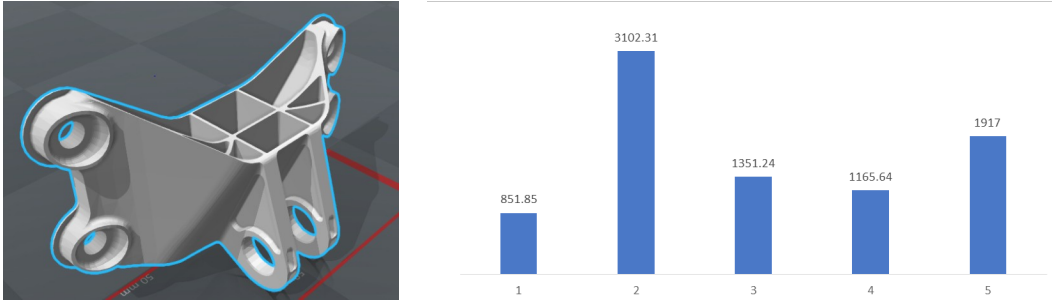

Figure 3-5: Aerospace engine bracket with price quotes obtained from various service bureaus (prices as of April 2018).

***Centralized Platforms***: Centralized manufacturing marketplaces on the other hand leverage information asymmetry across clients and service providers. Unlike decentralized platforms, centralized platforms determine the price, the service provider for the particular order and are generally responsible and liable for the parts fabricated by third party service providers. Innovations in applying data mining techniques has allowed these platforms to price parts through network enabled pricing. Rayna and Striukova [3] compare the pricing models of different platforms and show the wide variation in price mark-up charged by various platforms. They attribute the price variation to the relative immaturity of the market. We conducted a study to compare the price quoted by five different platforms for a heavily used 3D model – an aerospace engine bracket model made from aluminum alloy (AlSi10Mg). The price quote ranged from $850 to more than $3100 (Figure 3-1). With the level of price variation we found in the marketplace, it raises the question if these platforms are being fair to the suppliers and consumers. It becomes difficult for the designer to choose a platform to reliably procure 3D printing services. Conventional wisdom is to choose the platform with the cheapest price, however, trusting the quality from the cheapest source forces the consumer to shop around for more quotes while comparing process and material specifications from various sources. This again requires time and effort from the designer's end to evaluate price quotations from several service bureaus.

We propose an alternate mechanism for a platform that addresses the drawbacks of both centralized and decentralized platforms. Instead of having the service bureaus provide a price quote, we propose that the designers name their own price, a reverse auction type mechanism that procures additive manufacturing services from the service bureaus themselves. The designer bids for manufacturing services based on the design and the mechanism chooses a supplier which can match the bid requirements and stated price. Designers, especially price conscious designers, benefit by receiving the service at a price less than market value (price at which suppliers sell in the market) and suppliers obtain access to an alternate sales channel to sell their excess capacity to price conscious designers at a lower price (price lesser than the market value). Since the supplier is selling its extra capacity at a price lower than its market value, the identity of the supplier is hidden from the designer to prevent cannibalization of its existing sales until the client had placed the particular order. This mechanism has been successfully implemented in the travel industry by Priceline under the name "Name your Own Price™". The contribution of this chapter is to study if such a reverse auction mechanism creates desirable properties within a decentralized service system type marketplace.

The remainder of the chapter is organized as follows. In Section 3.2, we briefly review existing auction mechanisms and online auction scenarios such as those available in the hotel services, energy markets and cloud computing industry. Section 3.3 presents our novel mechanism. In Section 3.4 and 3.5, we present experimental simulation results to evaluate our model and analyze the impacts of various factors such as threshold price and supplier size on the results. Section 3.6 presents challenges and opportunities in this area and Section 3.7 draws conclusions and discusses ideas for further exploration.

## 3.2 Literature Review

Online auctions [4-8] and business models of the proposed type are not new. Goods and services are mostly sold using Posted Price, Negotiations and Auctions. With web enabled platforms rapidly giving rise to new business models, the scope of products which can be sold through online auctions has expanded [4]. The products sold through business to consumer online auctions mostly follow a mechanism similar to English auctions. English auctions augmented with an upper bound on price can increase the profit of sellers by increasing the expected bid from risk averse bidders. These auctions also increase the expected revenue of a seller or service provider as compared to the first price sealed bid – "Dutch bid" [6]. These auction types have been used in the Ebay like platforms [8], antique/precious arts industry, traveling the power distribution and more recently in the cloud computing business.

Power utilities and power generation companies use auctions to buy and sell electricity. Algorithms are written to maximize revenue of individual electricity suppliers and consumers in electricity spot markets [9-11]. Cloud computing uses online bidding for selling computing services. For example, Amazon Elastic Compute Cloud (EC2) [12] provides spot instances for which the users are required to bid for time. The spot prices fluctuate periodically based on the supply and demand for the spot instance. If the customer's bid value is greater than the spot price, the customer gets access to the computing resources until the task is finished or spot price becomes greater than the bid value [13-18]. Significant research has been conducted in this area to optimize users' bids and maximize suppliers' revenue. Zhang, Zhu and Boutaba [14] develop an optimal bidding strategy for Amazon EC2 cloud spot instance.

Online reverse auctions have been used in business to business industrial sourcing space as well as business to consumer space. One of the key industries to use such auctions is the travel industry where buyers bid for the services they wish to purchase and suppliers compete to match the bid. Priceline's 'Name Your Own Price™' model allows consumers to bid for travel services and the platform finds a supplier which can accept the price [19-21].

In recent work, Thekinen and Panchal [22] address the supplier selection problem in the area of additive manufacturing. They evaluate the application of existing match making mechanisms for optimal matching between service seekers and service suppliers. They however, do not consider price, a key factor in supplier selection, as a supplier and consumer attribute in their matching mechanism. To factor in supplier capacity, the work has assumed a limit on the number of orders a supplier can serve irrespective of order size. This is contrary to practices followed in the prototyping service bureau industry. Supplier capacities are limited by availability of machines/personnel and not by the number of orders. Our mechanism approach addresses these limitations on price and supplier capacities and we test our mechanism on a relatively large pool of suppliers with various types of machines and capacity level data.

In our model, designers propose a price that is acceptable to suppliers which reduces the inconvenience for a designer to haggle with multiple suppliers. It provides services to the designers, especially price conscious designers, at a price lower than the market value. It provides 3D Printing service bureaus an alternate sales channel to sell their excess capacity to price conscious designers. However, for such a marketplace to be sustainable, the selection mechanism must be fair to all three parties (the consumer, the bid platform agent and the service supplier). The main contribution of this work is that we have developed and evaluated a novel two stage algorithm that allows a designer to name their own price and have a qualified supplier willing to accept the stated price. This reverse auction type pricing model is a significant shift from existing service bureau marketplace practices. The approach also allows price of a particular service to be determined by the participants themselves – the client or the service provider and not by the middleware agent.

### 3.3 Reverse Auction Mechanism

The proposed mechanism consists of a network of independent suppliers who sell their services through a marketplace. Suppliers register their 3D printing machines on the marketplace. They provide parameters such as minimum order price, future machine availability, unit threshold price, available materials and process for each registered machine. Machines are registered on the platform as participants under the supplier that owns them to help match the technical requirements of the bid and match them to a corresponding machine. Key parameters are described in Table 3-1.

Let $S$ be a set of suppliers, $S = \{1, 2, \ldots, N\}$ and $s_i$, $i \in \{1, 2, \ldots, N\}$, represent the $i_{th}$ supplier. Let each supplier having up to $X$ machines. $j \in \{1, 2, \ldots, X\}$ represents $j_{th}$ machine of a supplier and $s_{ij}$ represents $j_{th}$ machine of $i_{th}$ supplier. Let $M = \{1, 2, \ldots, R\}$ represent a set of materials. Let $P = \{1, 2, \ldots, S\}$ represent a set of processes and $p_s$, $s \in \{1, 2, \ldots, S\}$ represent the $s_{th}$ process. Let $C_{ij1}, C_{ij2}\ldots\ldots\ldots C_{ij7}$ represent the availability of the $j_{th}$ machine of $i_{th}$ supplier for the next seven days. Let $thp_{ij}$ represents the unit threshold price (price per unit material volume) of $j_{th}$ machine of the $i_{th}$ supplier. Let $R = \{1.1, 1.2, 1.3 \ldots, 5\}$ represents the supplier ratings and $r_i$ represents the rating of $i_{th}$ supplier. Here we assume that the suppliers are able to price the utilization of the machine in dollars per unit volume of the part for a given machine.

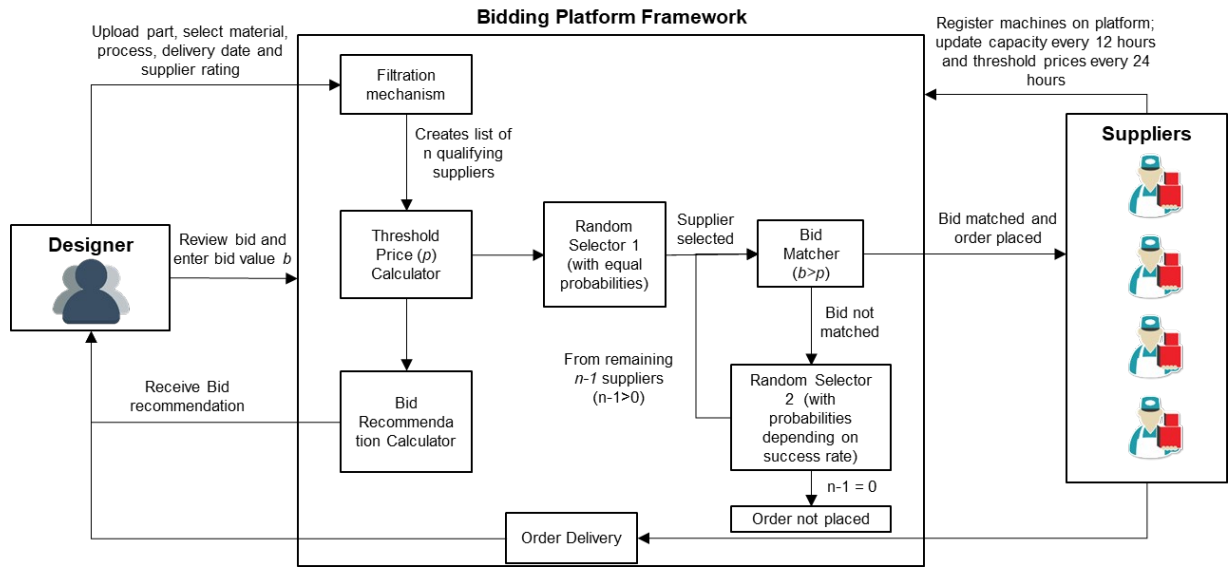

Figure 3-6: Reverse Auction Mechanism

Consumers or product designers begin by uploading the part design and choosing the preferred material, printing process and quantity. Let $\bar{P}$ be a part uploaded by a designer on the platform. Let $V_m$ be the material volume of the part which is the volume of material required to print the part. Let $t_s$ represents the approximate printing time of the part using the $s_{th}$ process. Our reverse auction mechanism (Figure 3-2) has the following modules:

***Filtration Mechanism:*** After the designer uploads the part $\bar{P}$ and specifies the desired material $m_k$, process $p_s$, delivery date, quantity and supplier rating $r$, the filtration mechanism creates a list $l_1$ consisting of $n$ qualifying suppliers with a total of $h$ qualifying machines capable of making the part. The mechanism works in the following steps. Step 1 filters those suppliers which have a supplier rating above the rating selected by the designer ($r_i > r$) and Step 2 matches the materials and processes provided by each filtered supplier with the material and process requested by the designer. Step 3 uses the approximate printing time $t_s$ of the part to identify the machines which have the availability to print the part before the desired delivery date ($C_{ij1} + C_{ij2} + \ldots\ldots > t_s$). The final list $l_1$ is the list of $n$ qualified suppliers with a total of $h$ qualifying machines which can create the part. Algorithm 1 (Figure 3-3) represents the functioning of Filtration Mechanism.

***Threshold Price Calculator:*** It calculates the threshold price $p$ for each of the $h$ qualifying machines filtered by the Filtration Mechanism. For each machine registered by the supplier on the platform, it needs to select a preferred method of determining the threshold price for a given part (based on either material volume or bounding volume or surface area or a combination of these). In such a case, it will provide the platform with a unit threshold price (assume, Dollars per unit material volume) which will be multiplied by the material volume of the designer's part to calculate the Threshold Price ($p = thp_{ij}*V_m$). Algorithm 2 (Figure 3-3) represents the functioning of Threshold Price Calculator.

***Bid Recommendation Calculator:*** It calculates a bid value to be recommended to the designer for part $\bar{P}$. This recommendation enables the designer to start a bid that has the highest chance of winning, particularly when bid attempts can be limited. There are two conditions which the recommended bid value should satisfy: 1) It should be less than the market value of the part for the platform to be competitive, and 2) There should be at least one supplier in the network which is able to match the recommended bid value (Figure 3-3, Algorithm 3). It recommends a threshold price between a lower bound *a* (minimum of threshold prices of *h* qualified machines) and an upper bound *b* (maximum of threshold prices of *h* qualified machines).

Table 3-4: Bidding Platform parameters

| Parameter | Description |
|---|---|
| Model | The 3D Model of the part to be printed, $\bar{P}$ with material volume, $V_m$. $t_s$ represents the approximate printing time of the part using the $s_{th}$ process |
| Bid | The price at which designer is willing to buy the service |
| Supplier Rating | This is based on the feedback from designers served by the supplier. It can vary on a scale of 1-5 with 1 being lowest rating and 5 being the highest. |
| Delivery Date | The date by which designer needs the product to be delivered at the desired destination. |
| Unit Threshold Price | Price at which the supplier is willing to sell the service for a machine in terms of $ per unit volume or any other metric. The suppliers need to frequently update the unit threshold prices to be competitive in winning orders. |
| Minimum Order Price | It is the price below which a supplier is not willing to accept an order. The supplier provides Minimum Order Price for every machine it registers on the network. |
| Availability | It is availability per day for the next 7 days of the 3D printing machine registered by the supplier, which is frequently updated. |

The recommended bid value has a direct impact on the profit margin of the bidding platform which consists of the difference between the bid value *b* entered by the designer and threshold price *p*. The designer is expected to either bid at the recommended value or a value lower than that. For instance, if the bid calculator recommends the lower bound *a* as the recommended bid value, the designer is expected to win the order at price *a*, which gives a minimal profit to the bidding platform but maximizes the designer's delight. However, if the system recommends the upper bound *b* to the designer, bidding platform has a chance to make a higher profit but may reduce customer satisfaction. We have defined a normalized index, termed as the 'Greediness Index', which describes the amount the bidding platform stands to gain based on the bid recommended price to the designer. Setting this index high automatically ensures that the bidding platform stands to gain the maximum margin, while a lower value will aid to the delight of the designer in being able to obtain a good deal. This index becomes important to entice designers to continue using the platform while giving the platform the flexibility to dynamically change the index on the fly.

The recommended bid is calculated using a power function $Y = a + (b-a)(I)^{1/2}$ where *Y* is the recommended bid value. The greediness index (*I*) is set by the bidding platform and can be based on the demand forecast of 3D printing parts. With a high demand forecast, Greediness Index (value between 0 and 1) can be set higher to gather more revenue for the bid platform, while a lower index can be dynamically set to entice designers to use the platform.

***Random Selector 1 and 2***: It randomly selects one of the *n* qualifying suppliers from list $l_1$ created by the Filtration Mechanism. With 'Random Selector 1', each of the *n* suppliers have equal probability of getting randomly selected. If the bid value is accepted by the supplier, the bid is won and the order is placed. In the event, a successful bid is not found in the first run, a 2$^{nd}$ round selection is made by 'Random Selector 2'. In the 2nd round, the probability of selection depends on the success rate (explained below) for each supplier.

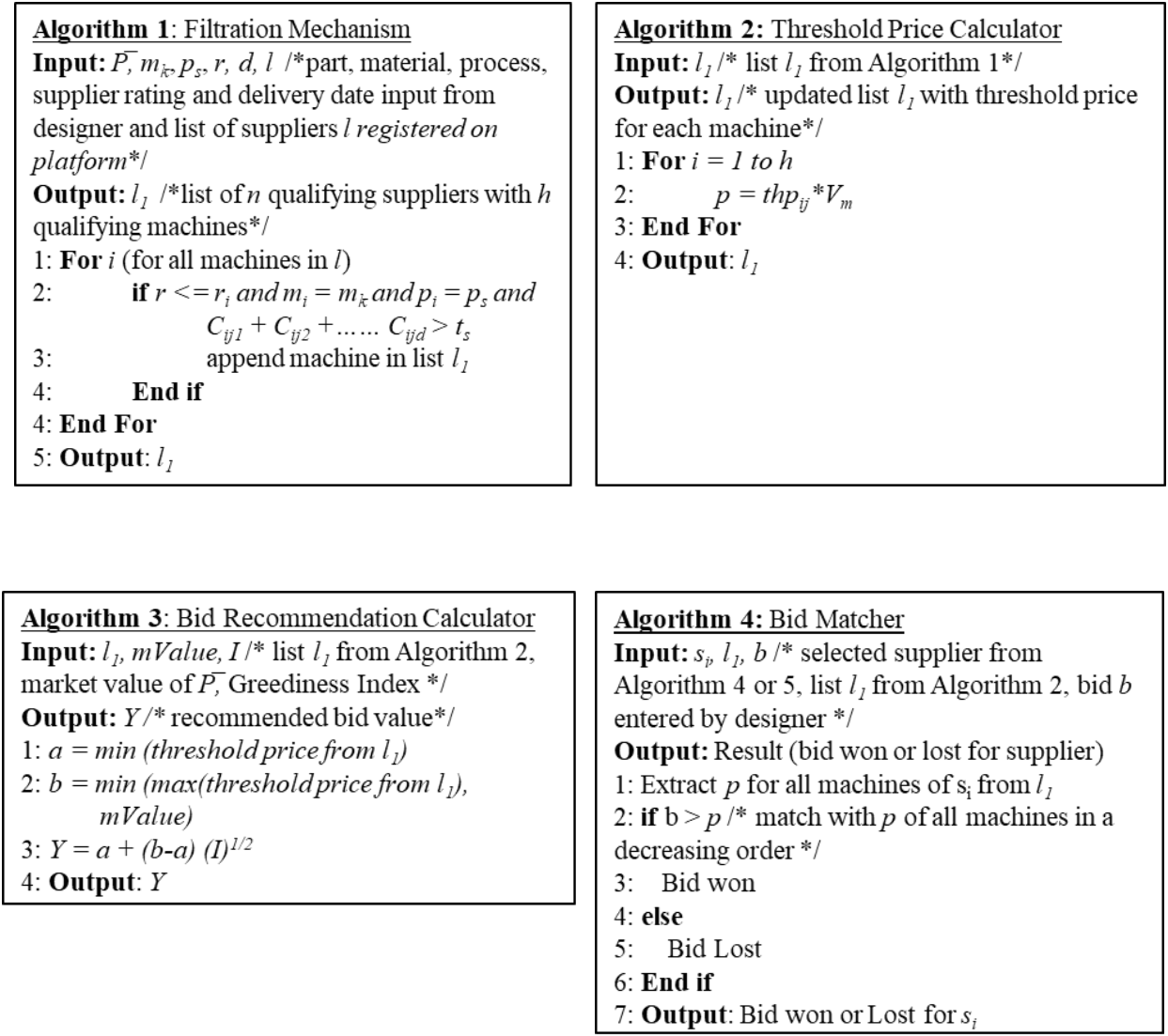

Figure 3-7: Core Algorithms Used in the Reverse Auction Mechanism

***Bid Matcher:*** It matches the bid value $b$ entered by the designer with the threshold price $p$ of part $\bar{P}$ for the selected machine $s_{ij}$. If the bid value $b$ is greater than the threshold price $p$ ($b>p$), the bid is won and order is placed (Figure 3-3, Algorithm 4). If none of the $h$ machines in the list $l_1$ of qualifying machines is able to match the bid, the bid is lost even after $h$ rounds of selection.

In the remaining part of this section, we will describe how the framework functions in its entirety. The designer uploads the CAD model, $\bar{P}$ on the bidding platform and selects the desired material, process, and minimum supplier rating and desired delivery date. The filtration mechanism creates a list $l_1$ of $n$ qualifying suppliers with their $h$ qualifying machines which can print the part $\bar{P}$. The Threshold Price Calculator calculates the threshold price of each of the $h$ qualifying machines in list $l_1$. The Bid Recommendation Calculator calculates the recommended bid and displays it to the designer. The designer reviews the recommended bid and enters a bid value $b$ which he/she is willing to pay. Random Selector 1 then randomly selects one of the $n$ qualifying suppliers. The bid matcher matches the threshold price of each qualifying machine of the selected supplier with the bid value. If the threshold price $p$ of any of the machines of selected supplier is less than the bid ($p < b$), the designer wins the bid and the order is placed. This is the first round of supplier selection.

If the threshold price of all the machines of selected supplier is greater than the bid ($p > b$), the bidding platform cannot accept the initial bid amount placed by the designer. In this case, the mechanism progresses to the second round of supplier selection and uses Random Selector 2 to select a supplier. It selects a supplier out of the remaining $n-1$ suppliers in list $l_1$. The probability of selection in this round depends on the success rate of each supplier in the first round. Success rate is the percentage of times a supplier's threshold price was greater than the designer's bid when it was randomly selected in the first round for all bids placed before the current bid. For example, a supplier which was able to meet the bid ($p < b$) 70% of the time when selected in the first round will have twice the probability of selection in the second round as compared to a supplier which was able to meet the bid ($p < b$) 35% of the times when selected in the first round. The success rate is calculated based on the selections only in the first round and not in subsequent rounds. After the supplier selection in the second round, the Bid Matcher matches the threshold price of the machines of the selected supplier with bid value $b$ entered by the designer. If the threshold price of any of the qualifying machines of this supplier is less than the bid ($p < b$), the designer wins the bid and the order is placed. This process is repeated until the bid is won or all of the suppliers are exhausted which results in the order not being placed.

If the bid is lost, the designer can rebid with a higher bid value up to n times (typically limited in a 24hr period). After $n$ bid attempts, the designer is not allowed to bid for the next 24 hours for the same attributes (design, material, process and supplier rating). Such limitations are placed to ensure that the bid platform is not gamed. If threshold price $p$ of more than one machine of selected supplier is less than the bid $b$, the algorithm chooses the machine with the highest threshold price. This provides the supplier with the freedom to print the part on any qualifying machine without making any loss. For instance, if the supplier has two low end fused deposition modeling (FDM) printers and both of them qualify for the order and have different threshold prices, it can print the part on any of the two machines. In such cases, the difference in the threshold price is not expected to be significant as both the machines are of same type.

This procedure of random selection ensures that every qualified supplier has a fair chance of getting selected (at least in the first round of supplier selection as the selection is random) and prevents any large supplier from gaming the system by selling at very low prices. The suppliers compete with the designer and not directly with each other which provides an incentive for small suppliers to remain on the platform. The mechanism promotes the suppliers to reveal their true threshold price since they do not directly compete with each other. The designer has limited attempts to game the system and is bound to reveal his true preferences in the limited number of bid attempts allowed by the mechanism. Figure 3-4 and 3-6 represent the reverse auction mechanism from a designer's and supplier's perspective and Figure 3-5 presents the reverse auction mechanism from bidding platform's perspective.

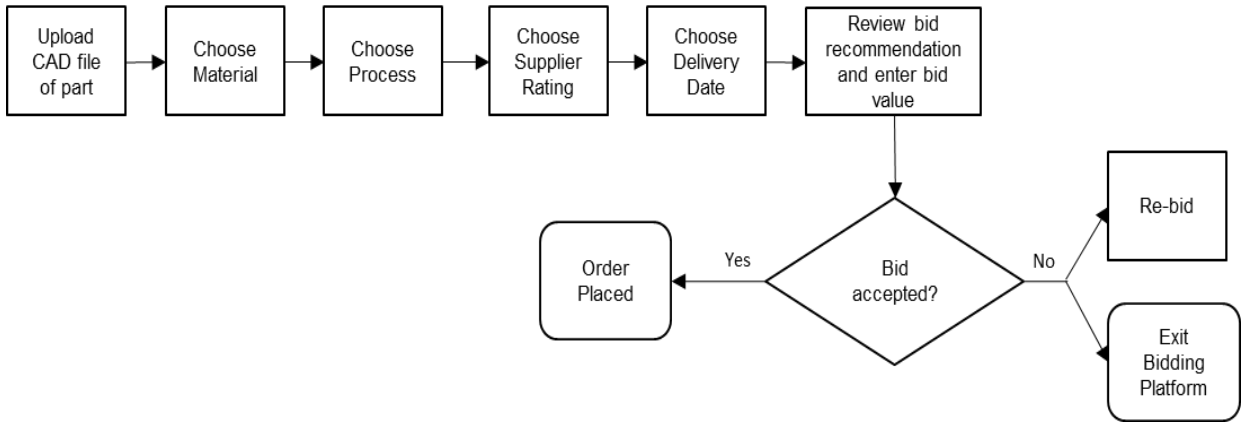

Figure 3-8: Reverse Auction Mechanism from a designer's perspective (Demand Side Platform)

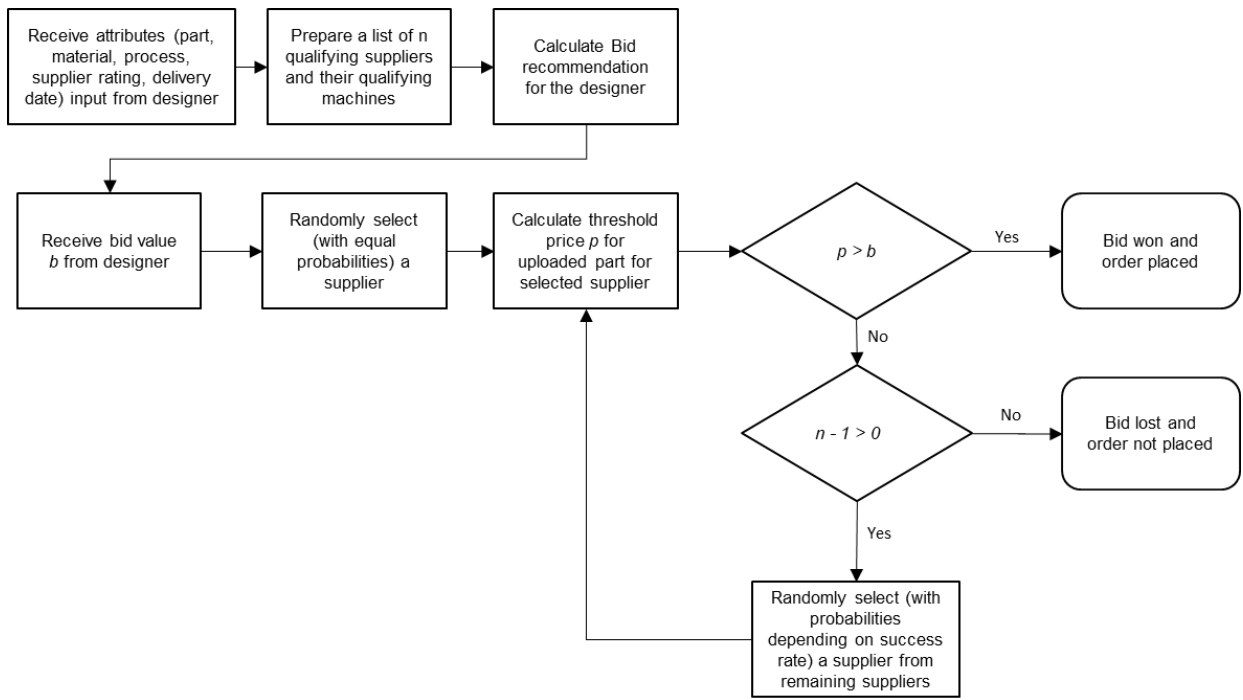

Figure 3-9: Reverse Auction Mechanism from Bidding Platform's Perspective

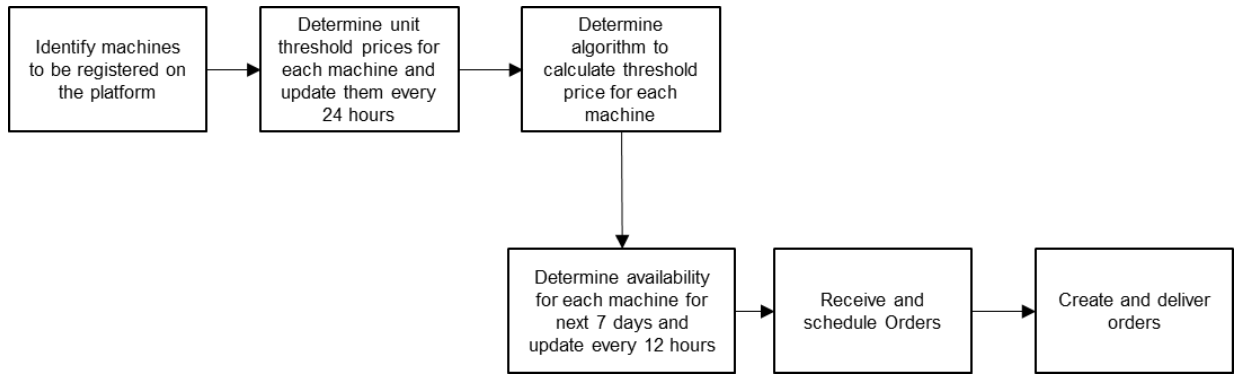

Figure 3-10: Reverse Auction Mechanism from a supplier's perspective (Supply Side Platform)

## 3.4 Simulation Model

The framework presented in the previous section was tested with a set of simulated data. We consider six different materials (from plastics to metal alloys) and a network of 125 suppliers with a total of 586 machines. Table 3-2 represents the processes and corresponding material assigned. Supplier ratings from 3 to 5 were considered with a resolution of 0.1. Though supplier ratings can vary on a scale of 1-5, we did not consider suppliers with a rating less than 3 as it is very uncommon for a designer to choose a supplier with a poor rating on a platform where an algorithm chooses a supplier for the designer. Designers require the order to be delivered within 4 business days from order date. We also assume that any part printed can be shipped overnight and 24 business hours are required for shipping and delivery. We consider a 16 hour working day (2 shifts of 8 hours each) for each supplier. We assume that machine availability for a supplier can be combined at the end or beginning of the day and the printer can be left working overnight. Suppliers update their machine availability status for each machine registered on the platform after an interval of 12 hours. Figure 3-7 represents sample designs used in the simulation model.

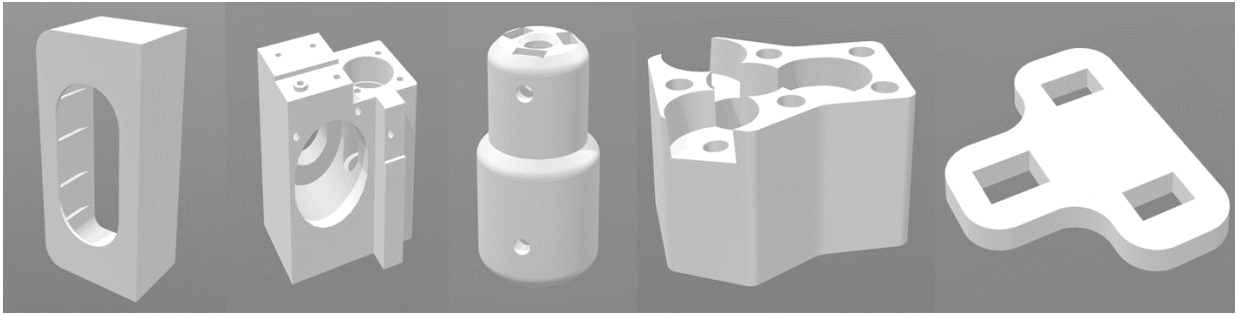

Figure 3-11: Sample designs used in the simulation model

Table 3-5: Processes and Materials used in experimental simulation

| Process | Material |
|---|---|
| Fused Deposition Modelling (FDM) | Polylactic Acid (PLA) |
| Digital Light Processing (DLP) | High Definition Acrylate (HDA) |
| Selective Laser Sintering (SLS) | Nylon |
| Selective Laser Melting (SLM) | Aluminum alloy |
| Direct Metal Laser Sintering (DMLS) | Stainless Steel Alloy (SS) |
| MultiJet Printing (MJP) | Wax |

Threshold price calculation algorithms were taken from the commercial marketplace 'Shapeways'. Threshold prices were based on parameters – Start-up cost, cost per unit material volume, cost per unit bounding box volume, cost per unit surface area and cost per unit machine volume. Start-up cost is a fixed cost component charged irrespective of the size of the part. Material volume is the volume of material required to print the part. Bounding box volume is the volume of a cuboid containing the part (length x breadth x height). Machine volume is the machine space occupied by the part. Threshold prices are represented in Table 3-3. Rows represent additive manufacturing processes used in our model and columns represent the parameters threshold prices are based on. Highlighted cell indicates that the parameter represented by the corresponding column was used in threshold price calculation for the process represented by corresponding row. For instance, fixed start-up cost and $/ material volume were used to calculate the threshold price for FDM, DLP, DMLS and MJP parts.

Initial success rate for each supplier has been considered to be equal and is updated after every bid attempt. Equal success rate in the beginning gives equal opportunity of winning an order to each supplier and as the simulations build up, success rates change, impacting the winning ability of suppliers in the second round. The threshold prices are calculated based on a fixed start-up price and unit price based on material volume, bounding volume and surface area of parts uploaded by the designers. Importantly, we vary threshold price with 3 factors: availability of machine, supplier rating and number of days between order and delivery date. These factors have been described below:

***Supplier Rating***: Suppliers with a lower supplier rating will participate in lesser number of bids as they will be filtered out for those customer bids which need a supplier with a higher supplier rating. Therefore, suppliers with a lower supplier rating would prefer a lower threshold price to increase their chances of winning orders in which they participate. We vary threshold prices with supplier ratings using this function:

$$p = a + (b-a) \frac{(r_i^{0.5} - r_{min}^{0.5})}{(r_{max}^{0.5} - r_{min}^{0.5})}$$

The supplier ratings $r_i$ varies between $r_{min}$ and $r_{max}$ and the threshold price $p$ varies between an upper bound $b$ and a lower bound $a$ set by the supplier for a specific machine.

Table 3-6: Threshold Price calculation Algorithms

| Processes \ Attributes | Start-up cost | $/ material volume | $/ bounding box volume | $/ surface area | $/ machine volume |
|---|---|---|---|---|---|
| Fused Deposition Modelling (FDM) | ■ | ■ | | | |
| Digital Light Processing (DLP) | ■ | ■ | | | |
| Selective Laser Sintering (SLS) | ■ | ■ | | | ■ |
| Selective Laser Melting (SLM) | ■ | ■ | ■ | ■ | |
| Direct Metal Laser Sintering (DMLS) | ■ | ■ | | | |
| MultiJet Printing (MJP) | ■ | ■ | | | |

*Machine Availability*: A supplier would keep the threshold price *p* of a machine lower if it has high machine availability and vice versa in order to keep the machines running as much as possible. We vary the threshold price of a machine [23] with capacity using the following function:

$$p = b + (b-a) C_{max}^{0.5} C_{min}^{0.5} \frac{\left(C\|ij\| - 0.5 - C_{min}^{-0.5}\right)}{\left(C_{max}^{0.5} - C_{min}^{0.5}\right)}$$

$C_{ij}$, varying between $C_{min}$ and $C_{max}$, is the average capacity of the $j_{th}$ machine of $i_{th}$ supplier for the days up to one day before delivery date. For instance – if the delivery date is three days from today, it will consider $C_{ij}$ as $(C_{ij1} + C_{ij2})/2$. Threshold price $p$ varies between the upper bound $b$ and lower bound $a$ specified by the supplier.

***Days before Delivery***: This quantifies urgency with which the designer needs the part to be printed and delivered. The supplier will charge more for an order which has to be delivered quickly. We vary the threshold price of a machine with 'Days before Delivery' $d$, varying between $d_{min}$ and $d_{max}$, using the following function:

$$p = b + (b-a) d_{max}^{0.5} d_{min}^{0.5} \frac{\left(d^{-0.5} - d_{min}^{-0.5}\right)}{\left(d_{max}^{0.5} - d_{min}^{0.5}\right)}$$

We vary $d$ between $d_{min}$ and $d_{max}$ and threshold price $p$ between the upper bound $b$ and lower bound $a$ specified by the supplier.

## 3.5 Results and Analysis

The simulation is intended to analyze the behavior of the model and impact of various factors such as supplier rating and threshold price on suppliers and designers. We track the following metrics during the simulation of our model. These metrics are core to the success of the business model.

1) ***Supplier's Delight***: It is the percentage of orders won by a supplier in a definite time period. The key factors which can impact the Supplier's Delight are Supplier Rating and Threshold Price of a supplier. To monitor the impact of the model on these metrics, we simulate our model for a week long period (8400 simulations considering an average of 600 order attempts every 12 hours). An order attempt signifies a designer bidding on the platform to procure a given 3D printing service. It can include multiple bid attempts of up to 5 bids and can result in either an order being placed or not being placed. Threshold prices were kept constant (without any variation with Machine Availability or Supplier Rating or Days Before Delivery) for all suppliers. Supplier's delight can be affected by primarily two factors – their supplier rating and the threshold price that they set for their excess capacity on a particular machine. We monitor the impact of these factors on the percentage of orders won by two largest suppliers 'Supplier 15' and 'Supplier 28' in our simulation model:

a) ***Supplier Rating***: Suppliers which meet the attributes (material, process, delivery date and supplier rating) desired by the designer qualify for an order. A rating 5 supplier will qualify for every order whereas a rating 3 supplier will qualify for least number of orders. Even if a supplier has a high success rate and lower threshold prices, its rating directly impacts its chances for qualifying for an order. Figure 3-8a displays number of orders won in a week by two largest suppliers when their supplier rating was varied. Threshold prices were not varied with Machine Availability or Supplier Rating or Days Before Delivery to monitor the impact of supplier rating. Our results validate that supplier rating has a direct impact on the percentage of orders won by a supplier even with random selection being used within the model. Varying threshold price with Machine Availability and Supplier Rating and Days Before Delivery in Figure 3-8b, we see combinatorial effect of supplier rating, threshold price and other factors. This indicates that even if a supplier has a low rating, it can win more orders than a supplier with a higher rating, if it can judiciously lower its unit threshold price to be competitive in the market.

b) *Unit Threshold Price*: Unit Threshold price has the highest impact on the number of orders won. In Figure 3-9, we can see the impact of threshold price on the number of orders won by 'Supplier 28'. It wins the maximum number of orders among all suppliers when it had the lowest threshold price among all. As the unit threshold price increases to average values, there is a significant drop in the number of orders won. When the threshold price is increased to the highest among all suppliers, it still is able to win a few orders. This also validates our rationale behind using random selection, since even with the highest threshold prices, a supplier has some chance of winning orders on the platform over the long term.

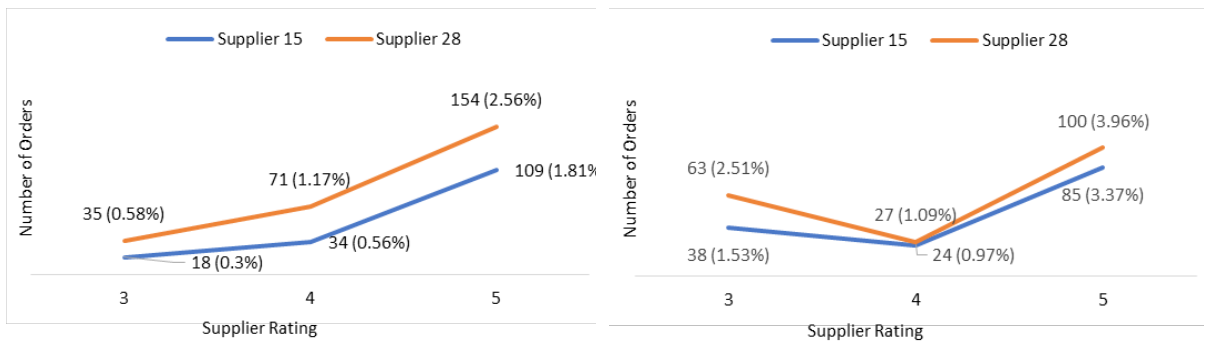

Figure 3-12: Percentage of orders won by two largest suppliers when Supplier Rating increase with a) fixed threshold prices and b) varying threshold prices.

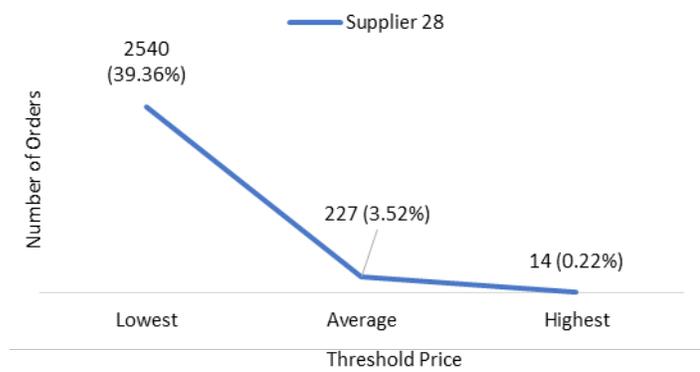

Figure 3-13: Number of orders won by 'Supplier 28' vs Unit Threshold Price

2) **Designer's Delight**: It is the percentage difference between order cost and market value of the product. Higher value signifies a better deal for the designer. We looked at how the average Designer's Delight varies with the Greediness Index ($I$) in Figure 3-10. It impacts the bid recommendation which further impacts the price at which the designer wins the order. As the greediness index increases, the bid recommendation value of the platform increases, which reduces the Designer's Delight.

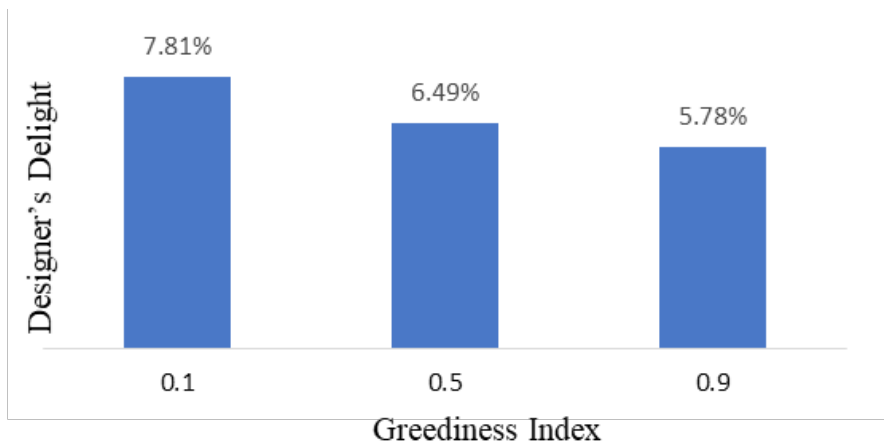

Figure 3-14: Designer's Delight vs Greediness Index ($I$)

3) ***Consistency***: The mechanism is consistent if entry or exit of suppliers does not change the dynamics of the platform. We consider a scenario in which a large supplier with 50 machines and a highest rating of 5 registers on the platform. When large suppliers enters a digital marketplace, they can tend to disrupt the ability of other smaller suppliers on the platform to win orders. Case 1 indicates a scenario when there are 125 suppliers in the system with suppliers 15 and 28 being the largest (12 machines each). Case 2 indicates a scenario when a new supplier is added with 50 machines. We find that adding this new supplier does not make a significant difference to any of the platform metrics (Table 3-4). The results show that the entry of a new large supplier in the system did not have any significant impact on the overall performance of the system. This is attributed to the random selection feature of the platform which ensures that every supplier has a fair chance of winning orders in the system.

Table 3-7: Performance metrics when a large supplier enters platform

| Parameter | Case 1 | Case 2 |
|---|---|---|
| Designer's Delight | 6.73% | 6.74% |
| Bidding Platform's Delight | 18.26 $ | 19.35 $ |
| Average Time Between Orders* | 12.76 hrs. | 13.09 hrs. |
| Successful Bid Attempts* | 30.1% | 29.8% |
| Number of suppliers winning at least one order | 125 (100%) | 126 (100%) |

*Successful Bid Attempts is the percentage of order attempts in which the bid was won and order placed. Average Time Between Orders is the mean of the average time between orders for suppliers which won more than one order.*

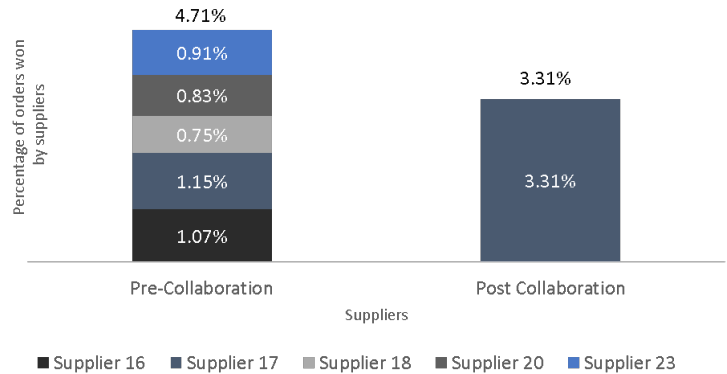

Figure 3-15: Percentage of orders won before and after collaboration by Suppliers 16, 17, 18, 20 and 23

4) ***Group Strategy Proof***: The mechanism is Group Strategy Proof if a coalition of suppliers misrepresenting information does not result in a better payoff for one or more suppliers in coalition. We consider a situation where a few suppliers collude and register together as a single supplier to increase the number of orders they win on the platform. To simulate this situation, we assume that suppliers 16, 17, 18, 20 and 23 collaborate and register as a single supplier. Supplier 16, 17 and 18 each have 7 machines and Supplier 20 and 23 each have 11 machines. Out of these suppliers, Supplier 17 has the highest supplier rating of 4.9. Hence, these suppliers collaborate and register as Supplier 17 with 43 machines to qualify for the maximum number of orders. The objective of suppliers 16, 18, 20 and 23 is to qualify for more number of orders by registering their machines through a supplier which has a higher supplier rating. The objective of Supplier 17 is to register large number of machines to win higher number of orders. The suppliers can have an agreement among themselves to share orders or profit margin which is not concerned with our model. We compare the results pre and post collaboration and find that post collaboration, the percentage of orders (3.31%) won by supplier 17 (a coalition of supplier 16, 17, 18, 20 and 23) decreases as compared to sum of percentage of orders won individually (4.71%) by these suppliers before collaboration (Figure 3-11). Since the mechanism chooses the suppliers randomly, members of a group of suppliers will have a higher probability of getting selected when registered individually as compared to registering as a single supplier. This indicates that the selected two-stage selection mechanism demotivates a potential coalition of suppliers which can misrepresent information on the platform.

5) *Fair Game*: It is the minimum number of bid attempts by designers which ensure at least one order for each supplier in the network. In a network of 125 suppliers with an average of 4.69 machines, it takes 6,000 bid attempts for each supplier to win at least one order on our platform. Figure 3-12 compares how the number of suppliers which won at least one order increases with the number of bid attempts for our platform than a directly competitive market place. In a directly competitive market place, suppliers directly compete with each other and the bidding platform chooses the supplier with the minimum threshold price. The results indicate that in a directly competitive marketplace, it takes larger number of bid attempts, more than 4 times as compared to our platform, for every supplier to win at least one order. The primary objective of using our algorithm is to ensure that every supplier has a fair chance of winning the game and there is no perceived direct competition among the suppliers (at least in the first round of supplier selection). This also ensures that suppliers are incentivized to provide a true valuation of their unit threshold prices for a healthy marketplace.

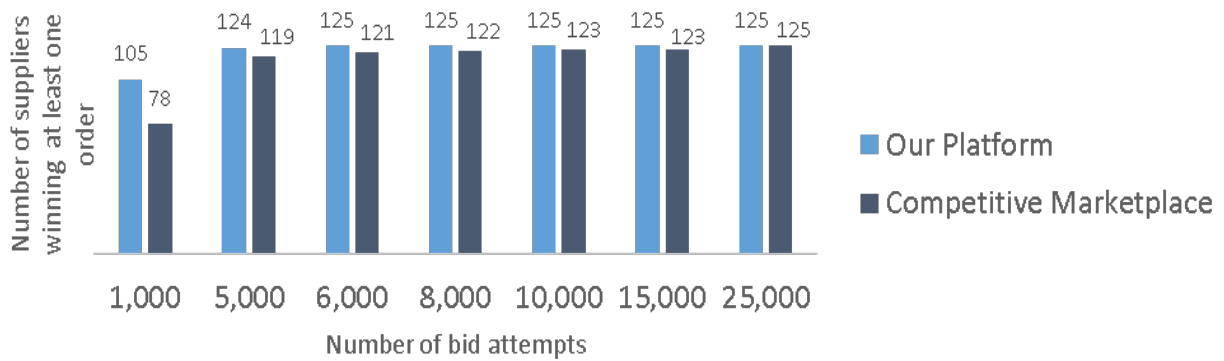

Figure 3-16: Number of suppliers winning at least one order on our platform vs. a directly competitive marketplace.

## 3.6 Discussion

Several challenges and research opportunities arise with such a mechanism for instant 3D printing services. Incentivizing suppliers to participate in the platform and having adequate number of end-user demand is necessary to maintain a healthy marketplace. Since suppliers are selling their available machine capacity at a price lower than the market value, the yield per machine, a measure of revenue generated by suppliers for the amount of inventory released to the platform has to be higher than the traditional way of soliciting business. The traditional way requires significant marketing campaigns to acquire and retain customers. Customer acquisition costs can be minimized by participating in such a marketplace. If a healthy marketplace is achieved, then we enable more 3D printing service providers, particularly those in remote places or those with limited budgets to participate in the prototyping industry for both 3D printing and traditional manufacturing one-off prototype production runs. Data privacy and protection will be key to ensuring that a supplier's business sensitive information such as threshold prices and pricing algorithms are not revealed to their competitors. Enabling excess capacity by service bureaus to be sold at lower rates can entice users to participate by obtaining cheaper orders without significant effort by either side in finding each other.

An instant check of machine capability to print a particular part is also a challenge. Currently there are limited robust algorithms which can verify the printability of a part on a particular brand of machine. New algorithms which can assess the status of any particular machine available on the platform against a model will further reduce barriers to the operation of such a marketplace. Supplier rating can be a sensitive issue. Current ratings as seen on 3Dhubs are determined by consumers who have used a particular suppliers services. New sophisticated methods maybe necessary to ensure that suppliers are not unfairly penalized from obtaining their fair share value from the platform. As 3D printing technologies transition to production scale, considering multiple unit quantities can be a significant feature added to such a marketplace. Reverse combinatorial auctions can be an interesting problem as it can serve as a point of negotiation between marketplace participants. More work must be carried out to diffuse enabling technology such as Machine-machine communication into the industrial base to leverage distributed intelligence and a digitally enabled supply chain for procuring parts.

## 3.7 Conclusion

We have proposed a new mechanism wherein designers bid with prices they are willing to pay and the mechanism finds a supplier which can match the stated bid price and order specifications. Such a mechanism automates the selection of the suppliers, saving time and effort on behalf of the designer. Suppliers can sell their excess capacity at a lower market price thereby gaining added revenue without significant customer acquisition costs. In addition, they need not reveal their true identity by having lower prices since it can affect regular sales. Maintaining equilibrium between the various entities in this business platform is key to ensuring the sustainability of the platform. Technologies such as instant price algorithms, automating machine availability reported to the platform and optimal matching of parts to supplier capability is critical to the success of the platform. New job scheduling algorithms capable of grouping several parts together in a bid can change how part cost and threshold prices operate [24]. Building in a recommended threshold price to be set by the supplier is essential to help balance supplier revenue and probability of winning bid orders. Ultimately, the proposed mechanism must be tested against real market data to truly understand supply and demand dynamics in a 3D printing prototyping marketplace.

# Chapter 4 : Stable Matching with Contracts for a Dynamic Two-Sided Manufacturing-as-a-Service (MaaS) Marketplace

Pahwa, D., Dur, U., & Starly, B. (2020). Stable Matching with Contracts for a Dynamic Two-Sided Manufacturing-as-a-Service (MaaS) Marketplace. (Under review in IISE Transactions)

## 4.1 Introduction

Two-Sided Manufacturing-as-a-Service (MaaS) marketplace platforms connect clients (e.g. product designer) requesting manufacturing services to suppliers providing those services. In the context of quick turn-around prototyping services and low quantity production runs, the platform has the ability to provide instant quotes and have the orders placed almost immediately. Clients obtain access to larger capacities and a variety of manufacturing capabilities compared to their traditional supplier network. On the other side, suppliers gain access to an additional source of revenue by selling their underutilized capacity. In a MaaS marketplace, the intermediary platform performs the task of order allocation and it is important that both types of participants find value in the allocation performed by the platform. If dissatisfied, the individual participants can reject the allocation and collaborate outside the platform or stop using it completely. Therefore, it is important for the platform to ensure that the order allocation or matching is stable [1]. A stable matching ensures that no two participants (clients / suppliers) will prefer each other over their assigned match. In other words, the participants receive their best available match given a set of suppliers and clients during a matching period. Research has shown that stable matches lead to a more successful marketplace [2] as the individual participants do not have incentive to bypass the platform which can lead to inefficiencies in the marketplace.

This work proposes matching mechanisms that enforce stability in order allocation in a MaaS marketplace ensuring that individual participants do not have incentive to bypass the marketplace making it sustainable in the long term. Order allocation is done by the platform while taking into account the preferences of clients and suppliers. Clients have preferences over supplier attributes such as preferring to work within a supplier's geographic location and suppliers have preferences over client attributes such as preferring jobs with certain material or client type. Such preferences are 1) Non-strict i.e. a client may be indifferent between two or more suppliers and vice versa; 2) Incomplete i.e. a client would prefer to remain unmatched over matching with a subset of suppliers and vice versa, and 3) Cardinal i.e. the utility received from a match can be quantified. Suppliers can accept multiple orders and their preferences over orders are interdependent. For example, a small order requiring 1 hour of production time can be one of the lowest preferences of a supplier which has an available capacity of 9 hours but it becomes important when combined with two other orders that require 8 hours of combined capacity.

Negotiations can play an important role in any MaaS marketplace. Clients can approach different suppliers and negotiate with them over attributes such as price or delivery date. Therefore, a client can have varying preferences over the same supplier depending on the attributes of the contract the supplier is willing to accept. This chapter considers matching with contracts framework [3] where participants on both sides have preferences over contracts. Additionally, the matching problem in a MaaS marketplace is dynamic in nature both clients and suppliers arriving at the marketplace stochastically. The match possibilities change with time in a dynamic setting and a myopic policy where agents are immediately matched on arrival might have short run benefits but significant disadvantages in the long run. Therefore, the platform may wait, termed as "Strategic Waiting" [4] for the market to thicken before allocating orders.

This work considers the bipartite matching problem in MaaS marketplaces in a dynamic environment where matching is performed repetitively over time and develops approximately stable matching solutions. Participants have non-strict, incomplete and interdependent preferences (for suppliers) over contracts on which the matching is performed. Mechanism design and mathematical programming approaches are used to develop these solutions. The approximately stable matching solutions are compared with a socially optimal matching to quantify the degradation in system performance due to enforcing stability. The contributions of this work can be summarized as follows:

- An approximately stable matching algorithm based on mechanism design and a maximum weight approximately stable matching solution using mathematical programming formulation are proposed to achieve stability in order allocation in a MaaS marketplace. A socially optimal (but unstable) matching is also formulated to compare its performance with approximately stable solutions. The comparison quantifies the degradation in matching performance of the marketplace by enforcing stability. It also evaluates the performance of all three solutions in terms of stability. Stable matching solutions are important for the success of a marketplace as demonstrated in other two-sided markets [2]. Unstable matches lead to participants bypassing the marketplace resulting in anarchy and inefficiency.
- The proposed solutions relax a key assumption considered in previous literature studying MaaS marketplaces where preferences of suppliers are specified over individual orders and are independent. In this work, suppliers are assumed to have interdependent preferences over groups of orders in which case a stable matching might not always exist. Additionally, matching is performed over contracts which enables negotiations between the participants on two sides over terms of the contracts. These considerations make the proposed solutions more viable since it takes into account realistic behavior in an actual marketplace. The proposed solutions are evaluated in a simulated MaaS platform under different assumptions on suppliers' and clients' behavior in the marketplace.

The remainder of the chapter is organized as follows. Section 4.2 presents the literature review, Sections 4.3 and 4.4 present problem description and proposed matching models respectively. Section 4.5 discusses the stability of proposed solutions. Section 4.6 describes the empirical simulation environment and Section 4.7 presents the results. Section 4.8 discusses ideas for further exploration and Section 4.9 draws conclusions from this work.

## 4.2 Literature Review

The concept of stability was first introduced in the stable marriage problem by Gale and Shapley [1]. It provides a stable solution in polynomial time and always finds a stable matching. It has been extended to applications such as matching doctors to medical residency programs [5, 6], school choice [7, 8] and others [9, 10]. Roth and Sotomayor [11] and Iwama and Miyazaki [12] provide detailed reviews of the literature on the stable marriage problem. Papers working on the stable marriage problem usually require the preferences of the participants on each side to be strict and complete. Gale and Sotomayor [13] showed that a stable matching can be found in polynomial time in case of incomplete preferences whereas Irving [14] showed that a weakly stable matching can be found in polynomial time under indifferent preferences. In a weakly stable matching, no two participants will strictly prefer each other over their assigned match. Under both, indifferent and incomplete preferences, Iwama et al. [15] show that a weakly stable matching exists for any matching instance.

In case of many to one matchings with indifferent, incomplete and interdependent preferences, as is the case with MaaS marketplaces, existence of a weakly stable matching is not guaranteed for all matching instances. Hence, it becomes important to find matchings which are close to stable [6]. However, there is also the possibility that multiple stable matchings can exist for a matching instance in a MaaS marketplace and finding the stable matching with the largest size (largest number of orders accepted) or the stable matching which maximizes a utility objective (collective welfare of all participants) is NP-hard [15]. Finding these specific stable matchings is an important consideration in a MaaS marketplace. Vande Vate [16], Rothblum [17], Roth et al. [18], Abeledo and Blum [19] and Vohra [20] study the stable marriage problem using linear programming approaches. Stable matching is characterized as a polytope whose extreme points are stable. Since there exist multiple stable matchings for any matching instance, mathematical programming approaches are helpful to find a stable matching which maximizes a utility objective. Matching with contracts [3], Hatfield and Kojima [21, 22] is a framework where agents on both sides have preferences over contracts rather than preferences over the agents on the other side. Sonmez and Switzer [23] apply matching with contracts to match cadets with branches in US Military Academy. They present the cadet branch matching problem as a special case of matching with contracts [3] model and demonstrate the application of the model beyond traditional two-sided matching problems.

Resource allocation methods in manufacturing industry focus on scheduling jobs to machines in a manufacturing facility with objectives such as maximizing efficiency, minimizing completion time or meeting due dates. Refer to Ouelhadj and Petrovic [24] and Liu et al. [25] for a survey on resource allocation and dynamic scheduling in manufacturing systems. This stream of literature does not involve the complexities associated with a network of independent suppliers. Thekinen and Panchal [26] study the application of several economic matching mechanisms in different manufacturing scenarios such as centralized / decentralized settings and determine the optimal mechanism for each setting. In later work, Thekinen et al. [27] determine the optimal frequency for matching orders with suppliers in a stochastic environment. However, they apply the deferred acceptance algorithm, to a manufacturing problem setting which liberally assumes that preferences of a supplier over orders are independent, do not match over contracts and do not maximize a utility objective. This is never the case in a manufacturing context.

## 4.3 Problem Description

A marketplace consisting of a network of independent suppliers on one side and clients requesting services on the other is considered. Suppliers register their machines on the marketplace platform in order to sell their underutilized capacity and clients in need of manufacturing services place orders on the marketplace. A client and his/her order is referred to as an order in the rest of this chapter. Orders, based on their attributes (such as required material, resolution and delivery date) are compatible with specific suppliers. The set of compatible orders and suppliers changes dynamically over time as new orders arrive and supplier capacities change. It is assumed that the platform has a planning period during which it waits for the market to thicken i.e. collect orders to allocate and then performs the matching at the end of the planning period. The allocated orders and supplier capacities leave the marketplace and new orders and additional capacities arrive in each period. Remaining orders and supplier capacities either wait for the next matching cycle or perish. The mechanism of order allocation is a key design parameter of the marketplace platform.

Let $S$ be a set of suppliers and $s_j, j \in \{1, 2, \ldots, n\}$ represents the $j_{th}$ supplier. Let $D$ be a set of orders and $d_i, i \in \{1, 2, \ldots, m\}$ represents $i_{th}$ order. Orders arrive in the marketplace with a mean rate $\lambda$ per period (assuming a Poisson process) and supplier reported capacities also arrive on the marketplace with a mean rate $\lambda_j$ hours per period (assuming a Poisson process) for $j_{th}$ supplier. Every supplier communicates their capacities for the next $q$ periods to the platform. An order specifies the following attributes: 3D design model to be fabricated, required material, due date, preferred 3D printing process and resolution required. A supplier's service listing must specify the following attributes: type of machine (3D printing process), available materials, printable resolutions and available capacities. A supplier can accept multiple orders depending on its available capacity, the production time required for the orders and their due dates. For simplicity, we assume that each supplier has a single machine, although the solution can be easily extended to suppliers with multiple machines. The marketplace platform accumulates orders to thicken the matching pool and allocates orders to suppliers at the end of every period.

Matching is performed by the platform based on the preferences of orders and suppliers over contracts. A contract $(i, j, c)$ consists of an order $i$, a supplier $j$ and terms $c$ such as material, 3D design, resolution, due date and price. An order supplier pair $(i, j)$ can have multiple contracts with different terms such as different prices, due dates or materials. Orders prefer certain attributes in contracts based on which it is determined the contract that has a higher preference over another. For instance, an order might prefer a contract with a supplier closer to its location which can deliver the order at a certain price and by a certain due date. Similarly, suppliers have preferences over terms of the contracts. For instance, a supplier might prefer a contract over another based on required material or customer type. Some of these attributes can be flexible i.e. negotiable, for example price, while some of them can be inflexible i.e. non-negotiable, for example 3D Design. Often there can be conflicting choices for an order or a supplier to make. For example a scenario where an order has to choose between a geographically closer supplier charging a higher price and a farther supplier charging a lower price. In scenarios where decisions have to be based on multiple criteria, expected utility theory, elaborated in Section 4.6.1, can be used to rank alternatives. It has been used for multiple criteria decision making [28-31] in a variety of problems such as supplier selection. The expected utility theory is used to quantify and rank the contracts in the order of preference. Each quantified contract assigns a utility value to the order and supplier.

4.3.1 **Stability in MaaS Marketplace**

A matching $M$ in a MaaS marketplace is a many to one allocation of contracts between orders and suppliers where each order is assigned to a single supplier through a contract and each supplier can be assigned to multiple contracts each related to a different order. A mechanism is a procedure of selecting a matching for each problem instance. The mechanism for matching in this chapter considers the notion of stability to allocate contracts. Stability for a MaaS marketplace is defined in two contexts. First, it is defined by the size of coalition of participants blocking the match. In this context, the concepts of *pairwise stability* and *group stability* are considered. An order supplier pair $(i,j)$ with an acceptable contract $(i,j,c)$ constitutes a blocking pair when it meets the following criteria: order $i$ is either unassigned or strictly prefers $(i,j,c)$ over its assigned contract in $M$ and supplier $j$ is either underutilized (i.e. it can accept $(i,j,c)$ in addition to the contracts assigned in M) or it strictly prefers $(i,j,c)$ over one or more of its assigned contracts in $M$. A matching in which there is no blocking pair is said to be pairwise stable. Under non-strict preferences, this definition of pairwise stability is termed weak stability by Irving [14]. The term pairwise stability refers to weak pairwise stability for the rest of this chapter. Group stability, a stronger concept of stability, ensures that a matching is not blocked by a coalition of participants of any size. In a MaaS marketplace, a group stable matching will ensure that there is no blocking coalition of size $n$ with 1 supplier and $(n-1)$ orders all of which would strictly prefer the coalition over their assigned matches. The coalition would consist of ($n-1$ contracts each having the same supplier and a different order. Since a stable matching is not always guaranteed for a matching instance (with incomplete, non-strict and interdependent preferences) in a MaaS marketplace, this work considers approximately stable matchings which maximize the degree of stability i.e. minimize the number of blocking groups in a matching. There can also exist multiple approximately stable matchings with same number of blocking groups but with different total utilities. Therefore, it is reasonable to seek an approximately stable matching with the highest total utility.

The second aspect affecting the concept of stability is the temporal dimension of matching. Since orders arrive on the platform continuously and supplier capacities change over time, the matching needs to be performed repetitively. In these dynamic settings, two notions of stability can be defined: transient and posterior stability. Transient stability is enforced in each matching cycle considering information (orders and supplier capacities) known in that cycle. However, new participants arriving in the next period can create blocking pairs/groups with participants in a transient stable match from the current period. When stability is enforced, considering information (orders and supplier capacities) received in two consecutive periods, it is called posterior stability. Posterior stability is achieved when participants in a matching cycle wait for an additional period for new participants to arrive and still fail to find a better match.

### 4.3.2 Impact of Stability

If the platform does not enforce stability i.e. elimination of blocking pairs/groups in the matching mechanism, it can find a matching which maximizes the total utility received by all participants. This matching is termed as maximum weight or socially optimal matching. Such a matching maximizes social welfare but might be unstable giving participants an incentive to deviate from their assigned matches and create blocking pairs/groups. Enforcing stability can prevent creation of blocking pairs/groups but it can result in degradation of total utility achieved. This degradation can be measured by comparing the stable and socially optimal matchings for a matching instance. As defined by Anshelevich et al. [32], the ratio of the highest and lowest total utility of the stable matchings to the total utility of the socially optimal matching is termed as price of stability and price of anarchy, respectively. The price of stability and price of anarchy quantify the loss of social welfare incurred by enforcing stability constraints in the system.

## 4.4 Proposed Matching Solutions

In this section, three formulations are proposed for establishing matching solutions in a MaaS marketplace. First, a maximum weight or socially optimal matching is formulated. This formulation maximizes the social welfare (collective value which all participants receive from the matching) and provides an upper bound on the total utility which can be achieved in a given matching instance. Second, a mathematical programming formulation that maximizes social welfare while enforcing stability (individual participants not having incentive to bypass the platform) is formulated and third, an approximately stable solution based on an extension of the Gale Shapley algorithm is proposed. Such an extension is known as a cumulative offer process [3].

### 4.4.1 Maximum Weight (MW)/ Socially Optimal Formulation

Consider a bipartite asymmetric edge-labeled multigraph $G$ with orders $i \in D$ and suppliers $j \in S$ as vertices. The contracts between feasible matches are represented as edges $(i,j,c)$ of $G$ where $c$ represents the terms of the contract between $i$ and $j$. Let $E$ be the set of all feasible contracts between all orders and suppliers in $G$. Let $X_{ijc}$ be a binary decision variable taking a value of 1 if the contract is accepted and 0 otherwise. Let $C_{ij} \subseteq E$ represents the set of contracts between order $i$ and supplier $j$ and $u_{ijc}$ the total utility gained by order $i$ and supplier $j$ from contract $c \in C_{ij}$. Note that here with slight abuse of notation $(i,j,c)$ and $c$ have been used interchangeably. $p_{ij}$ represents the production time in hours required to manufacture the order $i$ (due in period $q$) on supplier $j$'s machine and is same for all contracts in $C_{ij}$. $h_{jq}$ represents supplier $j$'s available cumulative capacity in hours up to period $q \in Q$ where $Q$ represents the set of periods in which the orders are due. Let $C_{ijq} \subseteq C_{ij}$ represents the set of contracts between order $i$ and supplier $j$ due on or before period $q$. The following matching formulation maximizes the total utility gained by all suppliers and orders in the platform.

$$\max \sum_{(i,j,c) \in E} u_{ijc} X_{ijc} \tag{1}$$

$$s.t. \quad \sum_{j \in S} \sum_{c \in C_{ij}} X_{ijc} \leq 1 \, \forall \, i \in D \tag{2}$$

$$\sum_{i \in D} \sum_{c \in C_{ijq}} p_{ij} X_{ijc} \leq h_{jq} \, \forall \, (j \in S, q \in Q) \tag{3}$$

$$X_{ijc} \in \{0,1\} \, \forall \, (i,j,c) \in E \tag{4}$$

Constraints (2) are matching constraints which ensures that a single contract is accepted for each order. Constraints (3) are the feasibility constraints which ensures that each supplier has the capacity to manufacture the accepted orders before they are due in period $q$. For each supplier and each period, these constraints ensure that the time required to manufacture the orders due up to that period is less than or equal to the cumulative capacity available with the supplier up to that period.

### 4.4.2 Maximum Weight Approximately Stable (MWAS) Formulation

This formulation finds a matching with the largest utility and highest degree of stability i.e. the minimum number of blocking groups. The graph $G$ in the MW formulation is modified as follows. A supplier can accept contracts from multiple orders based on its available capacity and therefore has preferences over combinations of contracts. Therefore, a new set of vertices $I$ representing combinations of individual orders $i$ are added to $G$. The graph $G'$ with the expanded set of vertices $D'$ where $D \subseteq D'$ contains all possible combinations of orders for each supplier. The vertices $S$ on the supplier side remain same as in $G$. With single order vertices $D$ in $G$, the number of edges (individual contracts $c \in C_{ij}$) between a supplier order pair $(i, j)$ is equal to the number of contracts between them. The number of edges (combinations of contracts $C \in C_{Ij}$) between a combined order vertex $I$ with $n$ orders and a supplier $j$ is $\binom{m}{1}^n$ provided there are $m$ contracts between each order $i \in I$ and supplier $j$. With the expanded set of vertices, an order $i$ can now be associated with multiple vertices in $D'$. $D_i \subseteq D'$ contains all vertices associated with order $i$. $(I, j, C) \in E'$ represents the set of edges in $G'$. $(I, j, C)$ is a unique combination of individual contracts $(i, j, c)$ between all $i \in I$ and supplier $j$ and its weight $u_{IjC}$ is equal to the sum of utilities gained by all orders $i \in I$ and supplier $j$ from these contracts.

Since suppliers have limited capacity, the size of the graph can be reduced by removing the infeasible nodes in $D'$ and infeasible edges in $E'$. First, all combined order vertices $I \in D'$ which are not feasible for a supplier are identified. Given a set of individual order vertices which are feasible for a supplier, the size of the largest feasible combined vertex is found by the following. $\max \sum_{i \in J} X_i$ subject to $\sum_{i \in J_q} p_{ij} X_i \leq h_{jq} \ \forall \ q \in Q$ where $i \in J$ denotes the set of individual order vertices for supplier $j$, $i \in J_q$ denotes the set of vertices in $J$ with due date upto period $q \in Q$ and $Q$ represents the set of periods in which the orders $i \in J$ are due. $p_{ij}$ represents the production time in hours required to manufacture the order $i$ on supplier $j$'s machine and $h_{jq}$ represents supplier $j$'s available cumulative capacity in hours up to period $q$ starting from the current period. $X_i$ represents a decision variable which is 1 if vertex $i$ is accepted and 0 if it is rejected. This formulation provides an upper bound on the number of orders a supplier can serve given a set of orders and its available capacities. It is solved for each supplier $j \in S$ and any nodes in $D'$ with size greater than the upper bound are removed from $G'$ for each supplier. Second, for the remaining nodes and edges in $G'$, feasible edges in $E'$ are identified using constraints (3) and infeasible edges are removed from $G'$. The MWAS formulation is executed in two steps. In the first step, a matching which minimizes the number of blocking groups is found as follows:

$$\min \sum_{I,j,C \in E'} Y_{IjC} \tag{5}$$

$$s.t. \sum_{j \in S} \sum_{C \in C_{Ij}} \sum_{I \in D_i} X_{IjC} \leq 1 \ \forall \ i \in D \tag{6}$$

$$\sum_{I \in D'} \sum_{C \in C_{Ij}} X_{IjC} \leq 1 \ \forall \ j \in S \tag{7}$$

$$Y_{IjC} + \sum_{i \in I} \sum_{I'j'C' \succeq_i IjC} X_{I'j'C'} + \sum_{I'jC' \succeq_j IjC} X_{I'jC'} + X_{IjC} \geq 1 \ \forall \ (I,j,C) \tag{8}$$

$$X_{IjC} \in \{0,1\}, Y_{IjC} \in \{0,1\} \ \forall \ (I,j,C) \tag{9}$$

Binary decision variables $X_{IjC}$ and $Y_{IjC}$ represent a combination of contracts $(I,j,C)$ between designers $i \in I$ and supplier $j$. $X_{IjC}$ is 1 if $(I,j,C)$ is accepted and 0 otherwise. $Y_{IjC}$ is 1 if $(I,j)$ forms a blocking group with combination of contracts $C$ and 0 otherwise. Constraints (6) ensure that each order is accepted only once and Constraints (7) ensure that each supplier accepts only one possible combination of contracts. Constraints (8) ensure for each $(I,j,C) \in E'$ that either edge $(I,j,C)$ is accepted $(X_{IjC}=1)$ or supplier $j$ is assigned another set of contracts $(I\|'jC')$ which it prefers at least as well (denoted by $\succcurlyeq$) as $(I,j,C$ or at least one of the orders $i \in I$ is assigned $(I',j,C'$ which it prefers at least as well as $(I,j,C)$. If none of the above conditions are met, then $(I,j)$ constitutes a blocking group with combination of contracts $C$ and $Y_{IjC} = 1$. In summary, either $(I,j,C)$ is accepted or at least one of the orders $i \in I$ or supplier $j$ is assigned a match which it prefers as well or $(I,j)$ becomes a blocking group with combination of contracts $C$. These constraints ensure that if there is a group (a set of contracts each with the same supplier and a different order) each of whose participant is either unmatched or strictly prefers being in the group to its assigned match, it will be considered a blocking group. The objective function in (5) minimizes the number of blocking groups and provides a lower bound (LB) on the number of blocking groups in a matching instance.

In the second step, the objective function is changed to $max \sum_{I,j,C \in E'} u_{IjC} X_{IjC}$ and a new constraint $\sum_{I,j,C \in E'} Y_{IjC} \leq LB$ is added in addition to constraints 6, 7 and 8. The formulation in the second step finds a matching with the highest utility among the matchings with the number of blocking groups restricted to lower bound found in the first step. The objective function in the second step can be modified to $min \sum_{I,j,C \in E'} u_{IjC} X_{IjC}$ to find a matching with minimum utility and to $max \sum_{i,j,C \in E'} X_{IjC}$ to find a matching with maximum cardinality. The minimum weight approximately stable matching provides a lower bound for the utility achieved and the maximum cardinality approximately stable matching maximizes the number of matches in any approximately stable matching.

### 4.4.3 **Approximately Stable (AS) Formulation**

This section proposes an extension of the Gale-Shapley algorithm. The graph $G$ in the maximum weight formulation (Section 4.1) is considered where we have vertices as the set of orders $i \in D$, suppliers $j \in S$ and edges $(i,j,c) \in E$. The algorithm executes as described in Figure 4-1. Each order $i \in D$ proposes to the supplier $j \in S$ in its highest ranked contract i.e. the contract with the largest order utility in its list of contracts. Since all edges $E$ in $G$ are feasible, the contracts are tentatively accepted by the suppliers. Note that even though the individual edges in $G$ are feasible, a supplier must evaluate which combination of edges/ contracts it can fulfill considering its capacity and the due date of contracts. This results in each supplier having a list of contracts accepted tentatively. The choice function (elaborated below), is then used to evaluate the tentatively accepted contracts for each supplier who then selects the subset it can fulfill while achieving highest utility. The remaining contracts are rejected. This completes the first round of tentative acceptance within a given period. The second round processes the orders in rejected contracts from the previous round. The rejected orders now propose to their second ranked contract and suppliers now choose from the combination of tentatively accepted orders from previous rounds and the current round. This process repeats until there are no rejected orders or all the contracts of rejected orders have been explored ensuring that the algorithm will always terminate for any matching instance.

The choice function chooses the set of contracts which can be processed by their due date and maximize the utility of a supplier $j \in S$. The maximum weight formulation reduced to a single supplier is used as a choice function. The objective function reduces to $max \sum_{(i,c) \in J} u_{j,c} X_{ic}$ where $u_{j,c}$ is the utility gained by supplier $j$ from a contract $(i,j,c)$, constraints (2) are no longer required and constraint (3) reduces to $\sum_{(i,c) \in B} p_{ij} X_{ic} \leq h_{jq} \ \forall \ q \in Q$ where $(i,c) \in J$ denotes the set of all contracts tentatively assigned to supplier $j$ with no two contracts having the same order $i$ and $(i,c) \in B$ denotes the set of contracts in $J$ with due date upto $q$. $h_{jq}$ represents supplier $j$'s available cumulative capacity in hours up to period $q \in Q$ and $Q$ represents the set of periods in which the tentatively assigned contracts are due.

---

**Approximately Stable Matching Algorithm**

**set** all orders as rejected orders

**set** an empty list for tentatively assigned contracts for each supplier

**input:** list of contracts for each order ranked by its utility

**while** any order $d$ is rejected **do**:

    **if** set of contracts for each rejected order is empty: **break**

    **tentatively assign** rejected orders to suppliers in their first ranked contract

    **add** assigned contracts to suppliers' list of tentatively assigned contracts

    **delete** first ranked contract from each order's list of contracts

    **for** each supplier $s$ **do**:

        **apply** choice function to accept a subset of contracts from tentatively assigned contracts

        **del** rejected contracts from its list of tentatively assigned contracts

        **append** orders from rejected contracts to list of rejected orders

    **end for**

**end while**

---

Figure 4-17: Algorithm for Approximately Stable Formulation

Though MWAS formulation will always find a stable matching if one exists, this algorithm (AS) may fail to find a stable matching the reasoning for which is explained as follows. If the choice function for suppliers satisfies a substitute condition, this algorithm yields matchings that are both pairwise and group stable as demonstrated by Hatfield and Milgrom (2005). The substitute condition states that any contract $(i,j,c)$ that is rejected by a supplier from a set $X$ is also rejected from any larger set $X'$ that contains $(i,j,c)$. The choice function described above fails the substitute condition because of interdependent preferences of suppliers over contracts in the MaaS marketplace problem setting. For instance, consider the following four contracts each with a different designer and same supplier with production time and supplier utility tuples as follows: $c_1$: (8.1, 0.95), $c_2$: (4.6, 0.80), $c_3$: (4.1, 0.72) and $c_2$: (4.4, 0.78). All the contracts are due in same period and supplier has a total capacity of 9 hours until the due date. From a smaller set of $(c_1, c_2)$, $c_2$ would be rejected. However, from any larger set of 3 or 4 contracts, $c_2$ will always be chosen. The impact of choice function failing the substitute condition would depend on the sequence of arrival of contracts in the algorithm. If $c_1$ comes first, it would result in violation of stability however, if it comes last, it would not impact the stability of matching. The stability/ existence of blocking pairs/groups in a matching instance can be arbitrary in this case. Therefore, in Section 7, the impact of the failure of this substitute condition on matching stability is evaluated empirically in a MaaS marketplace. Considering the marginal impact on stability found in results, this algorithm is termed as "Approximately Stable" for a MaaS marketplace problem setting. If suppliers specify predetermined number of acceptable contracts and preferences over contracts are independent, as assumed in Thekinen and Panchal, 2017, the AS algorithm described above becomes pairwise and group stable. However, the number of contracts acceptable to a supplier depends on the attributes of contracts and cannot be predetermined.

**4.5 Stability of the Proposed Matching Solutions**

4.5.1 **Transient Stability**

Recall that a blocking pair (BP) is formed by an order-supplier pair $(i,j)$ with terms of contract $c$ each of whom is either unassigned (underutilized in case of suppliers) or strictly prefers $(i,j,c)$ over its assigned contract(s). Let $u_{i,c}$ is the utility received by order $i$ and $u_{j,c}$ is the utility received by supplier $j$ from $(i,j,c)$ where $u_{ijc}=u_{i,c}+u_{j,c}$. For $(i,j)$ to qualify as a blocking pair with terms of contract $c$, $u_{i,c}>u_{i,c'}$ or $i$ should be unassigned and $\sum_{J'} u_{j,c} > \sum_{J} u_{j,c'}$ where $i$ has been matched to $(i,j',c')$, $J$ is the set of contracts to which $j$ has been matched and $J'$ which includes $(i,j,c)$ is the set of contracts chosen by the supplier from the set $J \cup (i,j,c)$. The set of contracts $J'$ are identified using the choice function described in Section 4.3. If $J \subset J'$ then the supplier $j$ is considered underutilized in the blocking pair $(i, j)$. If $i$ is unassigned, it is considered an unmatched participant in the blocking pair $(i, j)$. If the order in the blocking pair is unmatched and the supplier is underutilized, it is termed an available blocking pair. The MW matching and MWAS matching will always match an available blocking pair since it would always strictly increase the total utility. The AS matching can have an available blocking pair depending on the sequence of arrivals of contracts as described in Section 4.3. Similar to the procedure described for pairwise stability, blocking groups (BG) can be identified. For a group $(I,j)$ to be a blocking group, each $i \in I$ should have a feasible contract with $j$ and terms $c$ where $u_{i,c} \geq u_{i,c'}$ or $i$ should be unassigned and $\sum_{J'} u_{j,c} > \sum_{J} u_{j,c'}$ where $i \in I$ has been matched to some $(i,j',c')$, $J$ is the set of contracts to which $j$ has been matched and $J'$ is the blocking group $(I,j)$. Notice that this definition allows $J'$ to have some of the contracts assigned to $j$ in $J$. These contracts which are in $J \cap J'$ are indifferent between their current assignment and the blocking group. The contracts in the set $J'$ strictly prefer the blocking group over their assigned match. If $J \subset J'$ then the supplier $j$ is considered underutilized in the blocking group $(I,j)$. If an $i$ in $I$ is unassigned, it is considered an unmatched participant in the blocking group $(I, j)$. If all orders in the blocking group are unmatched and the supplier is underutilized, it is termed an available blocking group.

The MW matching and MWAS matching formulation will always match an available blocking group since it would always strictly increase the total utility. The AS matching formulation can have an available blocking group depending on the sequence of arrivals of contracts as described in Section 4.3.

### 4.5.2 Posterior Stability

Recall that posterior stability considers the orders and supplier capacities received during two consecutive matching periods. The procedure to identify the blocking pairs and groups remains same as in the case of transient stability with two key modifications. First, the set of participants which are considered to identify the blocking pairs is generally larger since participants from two consecutive periods are considered. Second, the choice function described in Section 4.3 is modified to maximize $\sum_{(i,c) \in J} u_{j,c} X_{ic}$ subject to $\sum_{(i,c) \in B_{tq}} p_{ij} X_{ic} \leq h_{tq} \ \forall \ q \in Q_t$, $\sum_{(i,c) \in B_{(t+1)q}} p_{ij} X_{ic} \leq h_{(t+1)q} \ \forall \ q \in Q_{t+1}$ and $\sum_{(i,c) \in B_q} p_{ij} X_{ic} \leq h_{tq} \ \forall \ q \in Q_t \cup Q_{t+1}$ where $t$ and $t+1$ denote the two consecutive periods in which orders arrive and $Q_t$ and $Q_{t+1}$ the set of periods in which the arriving orders in these periods are due respectively. $(i,c) \in B_{tq}$ represent all orders that arrive in period $t$ and are due in periods upto $q$, $(i,c) \in B_{(t+1)q}$ represent all orders that arrive in period $t+1$ and are due in periods upto $q$. $(i,c) \in B_q$ represent the set of orders that are due in periods upto $q$. $h_{tq}$ and $h_{(t+1)q}$ represent the cumulative capacity of supplier $j$ from period $t$ and $t+1$ to $q$ respectively. $(i,c) \in J$ represents all orders arrived in both periods $t$ and $t+1$.

### 4.6 Empirical Simulation Environment

The simulation environment consists of a marketplace with a network of 100 suppliers with each assumed to have one machine. The machines consist of 3D printers based on Fused

Deposition Modelling (FDM), Selective Laser Sintering (SLS), Stereolithography (SLA) and Material Jetting. The machine specifications include the materials and resolution at which it can process those materials. Materials include Polylactic Acid (PLA), Nylon, Polycarbonate (PC), Acrylonitrile styrene acrylate (ASA), Resins, Thermoplastic polyurethane (TPU), Aluminum and Steel alloy. Machines, resolution range and materials were taken from supplier profiles from the 3D Hubs marketplace dataset used in Pahwa and Starly [33] or from machine specifications available on vendor sites. Suppliers list their machine capacities of up to 6 hours per period on the platform for 4 periods into the future. Length of one period is considered to be 12 hours. The middleware platform has a reputation system where orders rate suppliers based on their service quality. Supplier ratings vary on a scale of 1 to 5 with 5 being the best rating. Though the 3D Hubs dataset had majority of supplier listings (84%) as FDM machines, we consider FDM having 50% of the listings, SLA 15%, material jetting 15%, polymer based SLS 15% and metal based SLS 5%. This provides a reasonable mix of all types of machines in a MaaS marketplace.

Orders that arrive on the marketplace consist of attributes such as 3D model information, material required, desired 3D printing process, required resolution and a due date. Different probability distributions are used to generate these attributes. For instance, Due dates for the orders vary from 3 periods up to 7 periods from the period in which the order arrived. The following probability distribution is used for generating due dates: 0.10, 0.15, 0.25, 0.25, 0.25 for 3, 4, 5, 6 and 7 periods respectively. 3D printing processes for the arriving orders have a probability mass function where the probability of FDM order arrival is 0.5, SLA 0.15, material jetting 0.15 and SLS 0.2 each. Average production time (production and post processing) required for an order arriving on the platform is 5.35 hours and orders have a Poisson arrival with a rate ($\lambda$) of 100 orders per period. An order is considered feasible for a supplier based on four machine attributes – 3D printing process, material, resolution and available capacity. Contracts are generated for each order and its feasible suppliers as represented by $(i,j,c) \in E$ (section 4.1). Additionally, feasible combinations of contracts are generated for suppliers as represented by $(i,j,c) \in E'$ (section 4.2). A contract consists of order attributes (material, process, resolution and due date), a supplier, and a price. Price is considered to be the flexible term in the contract and up to two contracts are generated between a supplier-order pair.

Suppliers value three attributes: the revenue to be obtained from the order, required material and urgency. The higher the revenue, the higher the utility for suppliers. The later the due date, the more time the supplier has for fulfillment, leading to higher utility. The utility from the material depends on the supplier's inventory. If the material is readily available with the supplier, the utility is higher. Orders value a supplier's size, rating, location and price to pay. Suppliers are categorized as small, medium or large scale and orders specify their preferences over supplier size. Some orders prefer to work with large firms whereas some prefer small scale workshops. Orders prefer monotonically decreasing price i.e. they would prefer a contract with a lower price to one with a higher price. There are five levels of supplier ratings with a higher rating meaning a better supplier and orders have a monotonically increasing preference for supplier rating. Location data for suppliers is obtained from the 3D Hubs dataset used in Pahwa and Starly [33]. Suppliers and orders are assigned a location (GPS coordinates) in the US and the distance between them is calculated. The closer a supplier is to an order, the higher its utility.

Incoming orders are accumulated in each period and matching is performed at the end of the period. Matched orders leave the marketplace and rejected orders remain in the platform to be considered in the next period until their due date. Supplier capacities are updated considering the assigned orders. New supplier capacities and new orders arrive in the next period. New orders along with the remaining orders from the previous periods are matched at the end of this period. This process repeats and the simulation is conducted over a length of 15 periods.

### 4.6.1 Quantification of Contracts

Consider an order $d_i$ and a feasible supplier $s_j$ with $C_{ij}$ representing the set of contracts between them. Expected Utility theory is used for quantification of contracts in this problem setting. The following steps are required to apply the expected utility theory to this problem. A detailed description of the steps followed is provided in Fernandez et al. (2005). From an order's perspective, first the important attributes of a contract are selected, the utility function for each attribute is assessed and the individual utility functions are then combined using a multi attribute utility function. For instance, order $d_i$ values $p$ attributes $a = \{a_1, a_2 \ldots \ldots a_p\}$ in its contracts and the corresponding utility functions of these attributes are defined by $u_{i1}(a_1), u_{i2}(a_2) \ldots \ldots u_{ip}(a_p)$. The multi utility attribute function is then defined by equation (10).

$$u_i(a) = f\left[u_{i1}(a_1), u_{i2}(a_2) \ldots \ldots u_{ip}(a_p)\right] \tag{10}$$

Assuming order's preferences for the attributes to be independent, an additive multi attribute utility function can be used resulting in equation (11).

$$u_i(a) = \sum_{r=1}^{p} w_{ir} * u_{ir}(a_r) \tag{11}$$

where $w_{ir}$ is the scaling factor associated with attribute $a_r$ for order $d_i$. Similarly, considering suppliers' attributes and individual utility functions, the multi attribute utility function for a supplier can be defined as in equation (12) where $w_{jr}$ is a scaling factor associated with attribute $b_r$ for supplier $s_j$ which values $q$ attributes with individual utility function for $r_{th}$ attribute denoted by $u_{jr}(b_r)$.

$$u_j(b) = \sum_{r=1}^{q} w_{jr} * u_{jr}(b_r) \tag{12}$$

The following represents an illustrative example of utility calculation for an order supplier pair $(d_1, s_2)$ over terms of contract $c$. $s_2$ operates on a large scale, has a rating of 3 and is located in Atlanta, GA. $d_1$ located in Raleigh, NC requires a 3D design manufactured with aluminum alloy at a resolution of 200 microns due in 4 periods. The individual utility function of $d_1$ for location is $u_{11}(a_1) = 0.595 * a_1^2 - 1.516 * a_1 + 0.925$ over a range ($50 \leq a_1 \leq 500$) miles, size is $u_{12}(a_2) = 1 \text{ if } a_2 = \text{'large'}, 0.6 \text{ if } a_2 = \text{'medium'}, 0.3 \text{ if } a_2 = \text{'small'}$, rating is $u_{13}(a_3) = -0.219 * a_3^2 + 1.225 * a_3 - 0.005$ over a range ($1 \leq a_3 \leq 5$) and price is $u_{14}(a_4) = 0.922 * a_4^2 - 1.962 * a_4 + 1.033$ over a range ($640 \leq a_4 \leq 880$). The multi attribute utility function for $d_1$ is $u_1(a) = 0.2 * u_{11}(a_1) + 0.1 * u_{12}(a_2) + 0.3 * u_{13}(a_3) + 0.4 * u_{14}(a_4)$. Given the attributes of the contract with supplier $s_2$ (distance between the location of supplier and order is 400 miles, price quote is 750\$ and the order prefers to work with a large scale supplier), the utility $u_{i,c}$ of $d_1$ over the contract is calculated as $u_{1,c} = 0.418$. For $s_1$, the individual utility function over material requirement of $d_1$ is $u_{11}(b_1) = 1 \text{ if } b_1 = \text{'aluminum'}, 0.7 \text{ if } b_1 = \text{'titanium'}, 0.3 \text{ if } b_1 = \text{'steel'}$, urgency of $d_1$ is $u_{12}(b_2) = -0.240 * b_2^2 + 1.329 * b_2 - 0.048$ over a range ($1 \leq b_2 \leq 8$) and revenue from the contract with $d_1$ is $u_{13}(b_3) = -0.444 * b_3^2 + 1.401 * b_3 + 0.032$ over a range ($150 \leq b_3 \leq 1600$). The multi attribute utility function for $s_1$ is $u_1(b) = 0.2 * u_{11}(b_1) + 0.3 * u_{12}(b_2) + 0.5 * u_{13}(b_3)$ and the utility

$u_{j,c}$ of $s_1$ over the contract is calculated as $u_{2,c}=0.611$. Similarly supplier and order utilities are calculated for all feasible contracts.

## 4.7 Results and Analysis

This section presents the results and analysis from simulation experiments conducted to evaluate the impact of stability in a MaaS marketplace. The matching solutions presented in Section 4 were implemented in using Python 3.7 and Gurobi 8.1. The experiments were performed on 3.70GHz Xeon CPU with 24 GB RAM. Following sections present the results from the simulation experiments.

### 4.7.1 Influence of Enforcing Stability on System Performance

The following metrics are used to quantify the impact of enforcing stability on platform's performance:

**Impact of stability:** This is the ratio of total utility achieved by stable matching to the total utility achieved by maximum weight matching. This represents the price of stability and anarchy for maximum and minimum weight approximately stable matching, respectively.

**Average participant utility:** This is the average utility gained by each matched participant evaluated separately for orders and suppliers. For suppliers, this represents average utility attained per matched order.

**Matched participants:** This is the proportion of matched participants evaluated separately for suppliers and orders.

**Average participant rank:** This is the average percentile rank of a matched contract evaluated separately for suppliers and orders.

Table 4-8: System performance metrics for proposed solutions with λ = 100; Averages over 5 instances of 15 periods each (95% confidence intervals (CI))

|  | AS | MWAS | MW |
|---|---|---|---|
| Impact of stability | 0.947 (0.007) | 0.954 (0.004) | 1 |
| Average order utility | 0.679 (0.003) | 0.681 (0.003) | 0.73 (0.003) |
| Average supplier utility | 0.681 (0.006) | 0.679 (0.006) | 0.676 (0.006) |
| Matched orders | 0.731 (0.016) | 0.738 (0.015) | 0.748 (0.015) |
| Matched suppliers | 1 | 1 | 1 |
| Average order rank | 0.282 (0.012) | 0.279 (0.009) | 0.234 (0.006) |
| Average supplier rank | 0.413 (0.026) | 0.422 (0.026) | 0.465 (0.016) |

Table 4-1 presents the results for the metrics defined above. In addition to the three proposed solutions, the minimum weight and maximum cardinality approximately stable matchings are also evaluated. The price of anarchy and price of stability are 0.917 and 0.954 respectively. This demonstrates that any approximately stable matching instance where the objective is not to maximize the utility will not be worse than 0.917 of the socially optimal. The AS matching with 0.947 impact of stability is very close to the MWAS matching. The results are in concurrence with Anshelevich et al. (2013) where they show that in case of asymmetric edge-labeled graphs, theoretically, the impact of stability can be arbitrarily bad. However, in simulated matching instances under different utility distributions, they show that the price of anarchy is above 0.85. Owing to a capacity constrained environment, all suppliers in the system are matched with at least one order in all three formulations. All three formulations and the maximum cardinality approximately stable matching which provides an upper bound on the number of matched participants also matches almost same number of participants (74% orders and 100% suppliers) establishing that in a capacity constrained environment all stable matches will assign almost same number of matches. Orders receive a higher utility in the MW matching leading to a lower average percentile rank compared to the other two solutions.

The computation time per matching instance (average over 15 periods) for AS matching is 9.4 seconds compared to 124.5 seconds for MWAS matching. The large number of stability constraints (equal to number of combinations of feasible contracts between suppliers and orders) in the MWAS formulation make it computationally expensive. For larger instances, the graph $G^{'}$ can be divided into subgraphs based on parameters such as machine types or location for computational efficiency. The subgraphs can still be large depending on the size of the network of suppliers in the platform. Algorithms such as Column Generation [34] can be developed to efficiently solve this formulation for larger instances. Since the AS matching performs equally well in terms of system performance and provides a significantly faster solution, it is recommended for solving larger instances of the problem.

### 4.7.2 Transient Stability of Proposed Solutions

The following metrics are used to evaluate transient stability of the solutions proposed in Section 4:

**Participants in blocking pairs/groups:** This is the number of unique participants in blocking pairs/groups divided by total number of participants evaluated separately for orders and suppliers.

**Average blocking pairs/groups per participant:** This is the number of blocking pairs/groups divided by number of unique participants in blocking pairs/groups evaluated separately for orders and suppliers. For suppliers, it is further averaged over the number of periods.

**Unmatched participants in blocking pairs/groups:** This is the number of unique unmatched (underutilized in case of suppliers) participants in blocking pairs/groups divided by the total number of unique participants in blocking pairs/groups evaluated separately for orders and suppliers.

**Available blocking pairs/groups:** This is the number of available blocking pairs/groups divided by the total number of blocking pairs/groups in a matching.

**Average participant utility gain:** This is the difference between the utility received from the blocking pair/group match and utility received from assigned match divided by the utility received from assigned match averaged over all participants and evaluated separately for orders and suppliers.

**Average size of a blocking group:** This represents the average number of participants in a blocking group. The number of participants in a blocking pair is always two. However, blocking groups can have a varied number of orders associated with a single supplier.

Table 4-2 presents the results for transient stability metrics. The AS matching has only a few participants in blocking pairs and the number of blocking pairs in which these participants are present are low. The performance is relatively worse in terms of group stability which is expected as it is a stronger notion of stability. The MWAS matching performs the best in terms of stability which is expected since this formulation minimizes the number of blocking groups. The MW matching is highly unstable as expected and performs the worst with almost half of orders and all suppliers in the blocking pairs and groups. These participants also have relatively

much higher number of opportunities to form matches outside the system compared to the other two solutions.

Table 4-9: Transient stability metrics for proposed solutions with λ = 100; Averages over 5 instances of 15 periods each (95% CI)

|  | AS | | MWAS | | MW | |
|---|---|---|---|---|---|---|
|  | **BP** | **BG** | **BP** | **BG** | **BP** | **BG** |
| Orders in BP/ BG | 0.006 (0.003) | 0.069 (0.011) | 0.001 (0.001) | 0.001 (0.002) | 0.44 (0.015) | 0.471 (0.014) |
| Suppliers in BP/ BG | 0.074 (0.054) | 0.312 (0.083) | 0.01 (0.018) | 0.01 (0.018) | 0.982 (0.02) | 0.984 (0.017) |
| Avg. BP/ BG per order | 1.025 (0.069) | 2.577 (0.435) | 0.4 (0.68) | 0.4 (0.68) | 2.84 (0.128) | 3.437 (0.306) |
| Avg. BP/ BG per supplier | 0.09 (0.018) | 0.289 (0.075) | 0.027 (0.045) | 0.027 (0.045) | 1.272 (0.08) | 1.424 (0.133) |
| Unmatched orders | 0.129 (0.174) | 0.326 (0.114) | 0 | 0 | 0.251 (0.033) | 0.241 (0.033) |
| Underutilized suppliers | 0.143 (0.113) | 0.03 (0.023) | 0.067 (0.185) | 0.067 (0.185) | 0.103 (0.065) | 0.103 (0.065) |
| Available BP/ BG | 0.079 (0.18) | 0 | 0 | 0 | 0 | 0 |
| Avg. order gain | 0.234 (0.19) | 0.454 (0.145) | 0.045 (0.08) | 0.032 (0.059) | 0.33 (0.025) | 0.3 (0.02) |
| Avg. supplier gain | 0.163 (0.132) | 0.416 (0.133) | 0.078 (0.205) | 0.06 (0.153) | 0.144 (0.013) | 0.147 (0.017) |
| Avg. size of BP/BG | 2 | 3.01 (0.023) | 2 | 0.967 (1.645) | 2 | 2.154 (0.036) |

### 4.7.3 Posterior Stability of Proposed Solutions

Table 4-3 presents the results when transient stable matches are evaluated for posterior stability. Participants in a specific period will have additional match opportunities owing to waiting participants from the previous period and if they choose to wait for the new participants arriving in the next period. Therefore, this represents the worst case scenario in a hypothetical situation where all participants will have up to thrice the match opportunities they receive in a transient stable matching. Considering this, period wise matching solutions are expected to perform significantly worse in terms of posterior stability as compared to transient stability. The results in Table 4-3 demonstrate that in terms of posterior stability as well, both AS and MWAS matching perform significantly better than the MW matching. Though all three matchings have almost equal number of participants in blocking pairs/groups, the number of opportunities to form blocking pairs/groups available to participants in AS and MWAS matchings are significantly lower compared to the MW matching. The posterior stability results demonstrate the importance of dynamic matching where agents can choose to wait expecting better matches in upcoming periods as discussed in Akbarpour et al. (2017).

Table 4-10: Posterior stability metrics for proposed solutions with $\lambda = 100$; Averages over 5 instances of 15 periods each (95% CI)

|  | AS | | MWAS | | MW | |
| --- | --- | --- | --- | --- | --- | --- |
|  | BP | BG | BP | BG | BP | BG |
| Orders in BP/ BG | 0.705 (0.031) | 0.891 (0.016) | 0.705 (0.022) | 0.888 (0.011) | 0.756 (0.012) | 0.938 (0.004) |
| Suppliers in BP/ BG | 1 | 1 | 1 | 1 | 0.998 (0.006) | 0.998 (0.006) |
| Avg. BP/ BG per | 4.007 | 51.574 | 4.039 | 44.215 | 7.16 | 105.074 |
| order | (0.145) | (15.85) | (0.208) | (11.896) | (0.218) | (24.474) |
| Avg. BP/ BG per supplier | 2.826 (0.287) | 19.71 (6.563) | 2.848 (0.305) | 17.565 (4.972) | 5.423 (0.333) | 43.115 (10.302) |

### 4.7.4 Influence of Switching Costs

The suppliers and orders in a blocking pair/group might not deviate from the match assigned by the platform if the benefit they are achieving by deviating is marginal. Also, to incentivize participants to accept its assigned match, the platform can impose a switching cost if a participant deviates from its assigned match. Therefore, it is important to analyze the benefit participants receive by forming blocking pairs/groups and deviating from the assignments in the proposed solutions.

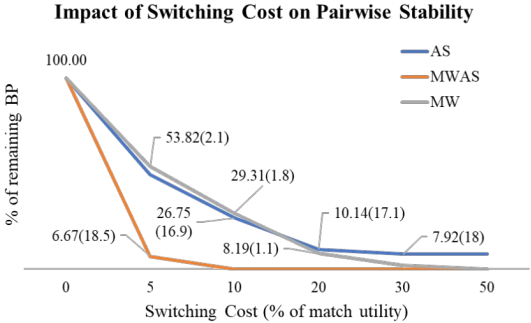

Figure 4-18: Impact of switching cost on pairwise stability; Averages over 5 instances of 15 periods each (95% CI)

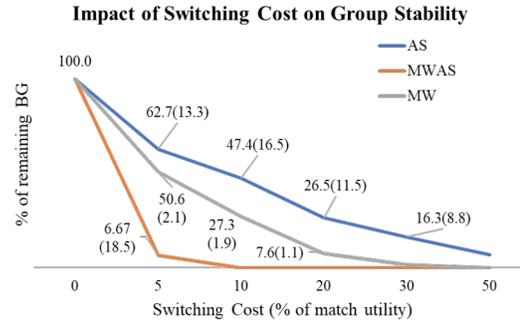

Figure 4-19: Impact of switching cost on group stability; Averages over 5 instances of 15 periods each (95% CI)

A multiplicative switching cost (Anshelevich et al., 2013) which is a proportion of the utility received by the participant in an assigned match is considered. Figure 4-2 and Figure 4-3 presents the impact of imposing switching costs on the number of blocking pairs and blocking groups for the proposed solutions. In the MW matching, the number of blocking pairs and blocking groups gradually decline as the switching cost increases reaching to almost zero at 30% switching cost. This demonstrates that the platform can enforce stability even with a MW matching by imposing a switching cost of at least 30% of achieved utility. The rate of decline of the blocking pairs and groups is almost same because the size of an average blocking group is 2.15 (from Table 4-2). The AS and MWAS matchings have significantly lesser number of blocking pairs/groups which varies highly among matching instances leading to wide confidence intervals for remaining blocking pairs/groups when the switching cost is increased gradually.

### 4.7.5 Impact of Unstable Matches over Time under Socially Optimal (MW) Allocation

In this section impact of unstable matches under the MW allocation is evaluated. When the matching is unstable, the participants can form matches outside the platform rejecting its allocation. Orders approach their known suppliers in the decreasing order of preference and if accepted, they reject the platform's allocation. To introduce this form of anarchy into the system, the first round of the AS algorithm is executed on the results of MW allocation with a key difference that these allocations of defying participants are not tentative but final. First a "Complete Access" scenario is considered where each order has access to the identities of all suppliers and second, a "Restricted Access" scenario is considered where orders only know the identities of suppliers they have transacted with in the past.

Figure 4-4 shows the impact of participants rejecting unstable matches on the total utility of the system. AS matching achieves 95.2% utility of the optimal allocation on average over a length of 100 periods whereas unstable matching under Complete Access achieves 74.16% of optimal utility. Under Restricted Access, the total utility remains close to the optimal allocation initially as fewer participants are able to reject the unstable matches. However, as the orders transact with more suppliers, their network of known suppliers grows and the total system utility almost reaches the level of complete access by the end of 100 periods. Figure 4-5 presents the percentage of defying participants under both scenarios. As the network of order grows under the restricted access scenario, the number of defying participants grows and almost reaches the level of defying participants under complete access. To prevent defying participants i.e. matches forming outside the platform, most, if not all MaaS platforms, hide the identity of the participants. These results demonstrate that the platform is better off over the long run being open and transparent. When the matching is stable, participants bypassing the platform is not a concern and the platform does not need to hide the identities and prevent access to the participants.

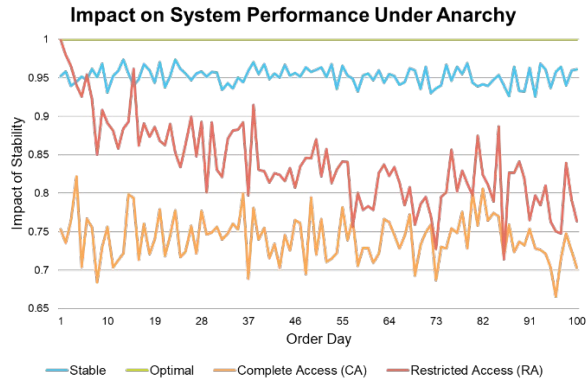

Figure 4-20: Impact on total utility when participants reject unstable matches

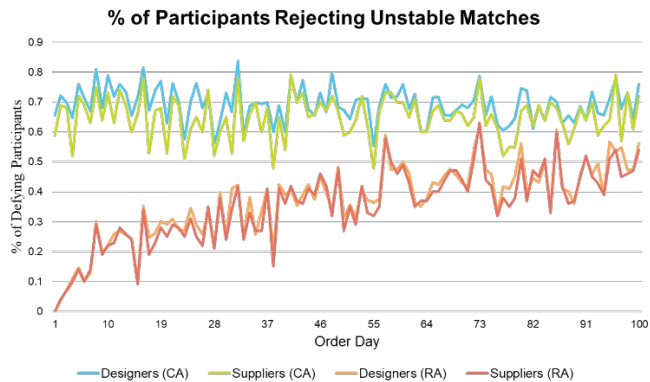

Figure 4-21: Percentage of participants rejecting unstable matches under restricted and complete access

### 4.7.6 Influence of Rate of Order Arrivals

In this section, the impact of rate of arrival of orders in the systems is studied. As the order arrival rate is increased, the suppliers receive larger set of opportunities to choose from and the system becomes more competitive for orders. This can be seen in the average percentile rank achieved by the suppliers and orders in the system (Table 4-4). As the order arrival rate increases, the average percentile rank becomes better for the suppliers and worse for the orders in both AS and MWAS solutions. The impact of stability slightly reduces as the order arrival rate increases demonstrating that as the system become more competitive or capacity constrained, the loss in utility due to enforcing stability increases. However, the AS matching and the MWAS matching perform quite closely at different arrival rates.

The impact of order arrival rate on the pairwise and group transient stability of the solutions was also evaluated. It is not presented here considering space restrictions. Since the number of blocking pairs and groups in the AS and MWAS solutions are few, the impact of order arrival rate is arbitrary. In the MW formulation, as the order arrival rate increases, the number of blocking pairs and groups increases. With increased number of orders in the system, participants have more opportunities to form blocking groups and pairs. The percentage of unmatched participants in blocking pairs and groups increases for orders and decreases for suppliers again owing to more number of orders for the same set of suppliers and their capacities in the system.

Table 4-11: Influence of order arrival rate (λ) on system performance; Averages over 5 instances of 15 periods each (95% CI)

|  | AS | | | MWAS | | | MW | | |
|---|---|---|---|---|---|---|---|---|---|
| λ | 80 | 100 | 120 | 80 | 100 | 120 | 80 | 100 | 120 |
| Impact of stability | 0.952 (0.002) | 0.947 (0.007) | 0.937 (0.006) | 0.96 (0.006) | 0.954 (0.004) | 0.944 (0.005) | 1 | 1 | 1 |
| Avg. order utility | 0.699 (0.012) | 0.679 (0.003) | 0.664 (0.008) | 0.698 (0.014) | 0.681 (0.003) | 0.669 (0.006) | 0.726 (0.004) | 0.73 (0.003) | 0.737 (0.002) |
| Avg. supplier utility | 0.659 (0.012) | 0.681 (0.006) | 0.698 (0.004) | 0.657 (0.012) | 0.679 (0.006) | 0.694 (0.003) | 0.662 (0.01) | 0.676 (0.006) | 0.685 (0.004) |
| Avg. allocated orders | 0.675 (0.019) | 0.731 (0.025) | 0.752 (0.021) | 0.682 (0.019) | 0.737 (0.025) | 0.757 (0.022) | 0.694 (0.02) | 0.748 (0.026) | 0.769 (0.02) |
| Matched orders | 0.844 (0.036) | 0.731 (0.016) | 0.63 (0.029) | 0.853 (0.036) | 0.738 (0.015) | 0.635 (0.028) | 0.869 (0.04) | 0.748 (0.015) | 0.645 (0.027) |

| Matched suppliers | 1 | 1 | 1 | 1 | 1 | 1 | 1 | 1 | 1 |
|---|---|---|---|---|---|---|---|---|---|
| Avg. order rank | 0.234 (0.028) | 0.282 (0.012) | 0.317 (0.011) | 0.237 (0.031) | 0.279 (0.009) | 0.308 (0.012) | 0.223 (0.013) | 0.234 (0.006) | 0.236 (0.005) |
| Avg. supplier rank | 0.493 (0.029) | 0.413 (0.026) | 0.339 (0.006) | 0.497 (0.028) | 0.422 (0.026) | 0.356 (0.012) | 0.518 (0.016) | 0.465 (0.016) | 0.414 (0.008) |

## 4.8 Discussion

Online manufacturing services marketplaces have emerged in several major markets globally with service abilities in short-run part production. However, as the demand increases, market grows with additional service offerings, leading to the need for enhanced transparency. Most current MaaS platform restrict the identities of the participants. Our analysis and simulation shows that in the long term, restricting the identity of the participants can adversely affect the platform health due to the lack of transparency and utility obtained by its participants when actively being within the platform. Sustaining operations of the platform can only be ensured when its participants do not bypass the platform. Stable bipartite mechanisms promote decentralization as the decision making is driven by the preferences of the participants. MaaS platforms can still be profitable when these marketplaces collect and store data on the suppliers and orders as related to collecting individual participants' preferences over time. Data-driven algorithms can be designed so that better contracts are automatically derived providing utility to both clients and suppliers. If such interaction value is provided, the participants do not necessarily have to by-pass the platform for conducting business.

On the technical side, computationally efficient approaches to solve matchings, particularly MWAS matching need to be developed. Additionally, matching frequency can impact the performance of the system as demonstrated by the results on posterior stability of matching algorithms. Participants can benefit from strategic decision making to enter into the matching pool or waiting for a better match [35] in each matching period. Such strategic decision making can lead to posterior stability in matching, which is equally important for long-term stability of the marketplace. The utilities of participants over feasible matches change with time and waiting for a match can also be expensive for participants. These additional dimensions provide avenues for further research in this area.

## 4.9 Conclusion

In decentralized MaaS marketplaces where manufacturing service requests from clients are assigned to suppliers by the marketplace platform, the participants (client and suppliers) can reject the assignment of the platform and find a better assignment on their own. Therefore, it is important for the platform to consider the notion of stability while matching the participants. This work introduces stable matching methods for a MaaS marketplace and empirically evaluates them for different criteria of stability. The results demonstrate that enforcing stability introduces a slight loss in the total utility achieved by the platform. However, it ensures a sustainable marketplace in the long run. Among the proposed stable solutions, the AS matching is recommended. It performs equally well in terms of utility, slightly worse in terms of stability but is significantly more time efficient compared to the MWAS matching. In a networked environment, MaaS marketplaces have an increasingly important role to play, particularly in prototyping service scenarios and its potential role in enabling the production of heavily customized limited quantity products.

# Chapter 5 : Dynamic Matching with Deep Reinforcement Learning for a Two-Sided Manufacturing-as-a-Service (MaaS) Marketplace

Pahwa, D., Starly, B. (2020). Dynamic Matching with Deep Reinforcement Learning for a Two-Sided Manufacturing-as-a-Service (MaaS) Marketplace. (under review in Manufacturing Letters)

**5.1 Introduction**

Two sided manufacturing-as-a-service (MaaS) [1] marketplaces connect clients requiring manufacturing services to suppliers providing those services. The platform removes friction in the manufacturing marketplace through quick decision-making such as providing instant quotations and order acceptance decisions. In the previous chapter we proposed a framework where the central platform performs the matching between orders and suppliers considering the preferences of both sides. The platforms collects the orders and available supplier capacities in each period and matches them at the end of each period. The approach is essentially a deterministic rolling horizon approach where the platform considers all the known information with in a period (planning horizon) and makes the decision at the end of the period. After the decisions are made (orders allocated and machine capacities reserved), planning horizon moves forward in time, collects new information (new orders and supplier capacities) and the process is repeated. The framework performs optimization at regularly spaced time intervals. In this chapter we propose a framework with two key modifications. First, instead of the platform making the matching decisions, suppliers themselves accept or reject the incoming orders. As discussed in Chapter 1, in a truly decentralized marketplace, participants take their own decisions and the platform supports them with decision making tools.

Second, suppliers are interested in maximizing the value they derive from the available capacity over time. Orders and machine capacities arrive stochastically over time and suppliers need to make decisions on whether to accept or reject the orders. They have limited resource availability i.e. machine capacity and they face the task of maximizing the value i.e. revenue they derive from the available resources over time. This problem is essentially a dynamic stochastic knapsack problem (DKSP) [2]. The rolling horizon framework proposed in the previous chapter [3] enables slightly better optimization than a dispatching rule as the orders and suppliers are batched up to optimize allocation in a single period. However, that approach is still myopic as it does not consider the future arrivals of orders and supplier capacities. We demonstrated this limitation by evaluating the posterior stability of approximately transient stable solutions. The posterior stability was significantly worse than the transient stability primarily because of new match possibilities arising from new orders and supplier capacities arriving on marketplace over time. Since the manufacturing marketplace is highly dynamic, it is important to consider the impact of future arrivals on the current matching decisions.

This chapter models this sequential decision making problem for suppliers with a Markov Decision Process (MDP) framework and uses reinforcement learning to solve it. By considering the currently available orders, expected future orders, and its current and expected future capacity, the supplier should be able to refine its strategy to better allocate its available capacity to the incoming orders. This is especially useful in scenarios when the available capacity is limited compared to demand in the marketplace. In a marketplace setting, with limited capacities, it is important to devise a sequential order acceptance policy so that the available capacity can be utilized for accepting orders considering both the immediate and future rewards.

In a marketplace environment, the distributions of order arrivals and the incoming supplier capacities can change over time. Therefore, a learning approach where an agent can learn the optimal policy and automatically adapt to a changing environment is considered. A model based reinforcement learning approach such as dynamic programming is not preferred as the transition dynamics are hard to capture in such an environment. Traditional reinforcement learning methods [4] cannot handle large state spaces. Function approximation with neural networks in Deep Reinforcement Learning (DRL) solves the challenge of computational intractability. Moreover, this framework can be implemented in a distributed manner where each supplier solves its own MDP and learns from experience in the environment. Since the number of possible allocations increase exponentially as the number of suppliers and orders in the marketplace increase, using a centralized DRL agent is not feasible. In this chapter, we consider this problem for a single supplier to establish the validity of the approach and the problem in multi agent settings is left for future work.

**5.2 Deep Reinforcement Learning**

Reinforcement learning (RL) [4] is a method where an agent learns to map situations in an environment to actions in order to maximize a numerical reward signal. The key elements of a reinforcement learning system are a policy, a reward signal, a value function and an optional model of the environment. The policy determines what action to take in which state and the agent learns the policy by acting in the environment. The reward signal defines the objective of the agent. At each time step in the environment, the agent takes an action and receives a reward. The agent's objective is to maximize the reward in the long run. The reward signal defines for the agent which actions are good and which are bad in order to meet its objective. The reward signal indicates the immediate reward for the agent i.e. the benefit it receives immediately after taking an action. The value function determines what is good for the agent in the long run. The model of the environment determines the future state of the environment after an action is taken and is used for planning the course of action. The methods which use models for planning are called model based methods whereas methods which do not require a model of the environment and solely learn from trial and error are called model free methods. Figure 5-1 represents the agent environment interaction in an RL framework.

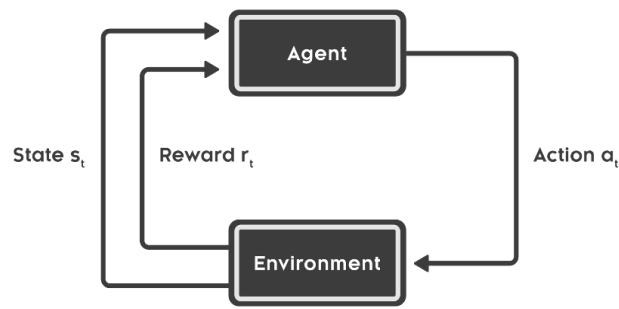

Figure 5-22: Agent environment interaction in a reinforcement learning framework

In a conventional reinforcement learning framework for example Q-learning, a lookup table is used to store the value of agent for each state $s_t$, action $a_t$ pair. However, the size of the table exponentially increases with the number of features representing the state. This phenomenon is called curse of dimensionality and it makes the table computationally intractable. Also, the method leads to training data sparsity i.e. some states might not be visited by the agent enough for it to train on those states and converge. When the number of states is large, neural networks can be used for approximating the value function for the state making it computationally tractable. Additionally, function approximators are generalizable. For the states which are not visited enough, they can identify similar states and use them for training instead.

Deep neural networks for function approximation combined with reinforcement learning is termed as Deep Reinforcement Learning (DRL). DRL methods have been recently applied to sequential control applications such as automated agents playing computer games [5] and sequential decision making in a variety of marketplace applications which have been discussed in the next section.

## 5.3 Literature Review

Previous works consider resource allocation as a static optimization problem [6, 7] which is not suitable for a dynamic MaaS marketplace. Dynamic methods such as rule based heuristics or rolling horizon approach [8, 9] have also been widely proposed, however, they can also be myopic in nature or require reasonably accurate forecasts of future arrivals. Additionally, most dynamic scheduling/ dispatching methods [9] have focused on scheduling jobs on machines in a manufacturing unit. This stream of literature does not involve the complexities associated with a network of independent suppliers. In a MaaS marketplace, suppliers need to accept or reject the orders fairly quickly and scheduling/ batching decisions can be taken later considering orders from the marketplace and other sources. In manufacturing systems with multiple suppliers, centralized approaches [10] that consider all participants (suppliers and orders) for allocation may not result in choosing an optimal action for each supplier individually. With suppliers having different preferences and objectives, decentralized [11] decision-making is desirable to maximize participation within a MaaS platform.

Anshelevich et al. [12] determine a threshold where an agent with a match utility less than the threshold would reject the match and wait for a better match in the future. In a MaaS marketplace, the environment is non-stationary and the threshold will change over time. Therefore, a learning approach which can adapt to the changing environment is preferred over determining a static threshold which would need to be reevaluated over time. Moreover, thresholds for each supplier can vary with different categories of orders making the method infeasible to apply in a large marketplace with thousands of suppliers.

DRL has been used in sequential decision making for resource allocation in a variety of marketplace applications such as ride sharing [13-16], display advertising [16, 17] and cloud computing [18, 19]. Ride sharing application also comprises of a two sided marketplace where riders need to be matched with drivers. Agents in these marketplaces also have a spatial dimension (the physical location of vehicles and riders) in addition to the temporal dimension which makes the state space even larger. DRL methods have been extensively studied in the ride sharing literature to assign riders to drivers (similar to assigning orders to suppliers in a MaaS marketplace) or for repositioning vehicles to an area with higher future demand. Lin et al. [13] design a contextual multi-agent reinforcement learning framework and propose contextual deep Q-learning and contextual multi-agent actor-critic algorithms for fleet management via coordination among driver agents. Al-Abbasi et al. [14] considers vehicle repositioning in a ride sharing environment where multiple passengers share a ride and uses deep Q-network (DQN) to determine an optimal policy. They propose dispatching using a distributed approach where each driver agent can take its own dispatching decisions. Jintao et al. [15] considers ride request as an agent and determines optimal time to enter the ride sharing pool using deep Q-network and actor critic network. Qin et al. [16] discusses how the ride share platform (DiDi) in China has evolved from using combinatorial optimization based approaches to DRL approaches for dispatching decisions.

Another key application area in literature for DRL is display advertising where advertisers need to determine the bid for an advertising impression under a constraint budget.

The advertisers pay a certain cost for the impression and receive a certain value from the impression space measured with click through rates or other metrics such as revenue accrued from sales. The objective of advertisers is to bid on impressions which have high returns and minimal costs over the duration of the ad campaign under the given budget. In a MaaS marketplace, manufacturers face a somewhat similar decision while accepting the orders under a capacity constrained environment. They receive a certain value (measured in terms of revenue or any other metric) from an order and pay a certain cost (in terms of capacity consumption) and their objective is to maximize the value they derive from the available capacity in the long run. Cai et al. [17] uses a model based reinforcement learning method to determine an optimal bid for an advertiser under a constrained budget in a display advertising application. They use neural networks to approximate the value function and dynamic programming to solve the problem. Wu et al. [18] uses a model free approach (DQN) to solve the budget constraint bidding problem in display advertising. Another application area where DRL methods have been studied is scheduling in cloud computing. Liu et al. [19] uses DQN and Mao et al. [20] uses policy gradient method for resource allocation in cloud computing environment. The successful application of DRL in these marketplace environments is the primary motivation to test the use of deep reinforcement learning for matching in manufacturing marketplaces. All of the above marketplace environments share the common goal of sequential decision making under constrained resources. Conventional approaches such as dispatching heuristics and combinatorial optimization methods result in myopic solutions. Moreover, the combinatorial optimization methods are NP Hard making them inefficient for marketplace applications which requires quick online decision making.

## 5.4 Methodology

Figure 5-2 describes the interaction between supplier agent and the environment. A MaaS marketplace is considered where orders arrive with a rate $\lambda_d$ per period. An order consists of attributes such as 3D design, material requirement and due date. The supplier commits capacity to the platform for the next $q$ periods. Capacity arrives randomly over time with a Poisson distribution parametrized with mean rate $\lambda_s$ hours per period. Material availability of the supplier also changes randomly over time. The supplier quotes a price for the order depending on the utility it would derive from it. The utility depends on the order's attributes and supplier's available capacity and materials. The objective of the supplier is to maximize its revenue over a period of time T. The MDP is formulated as follows:

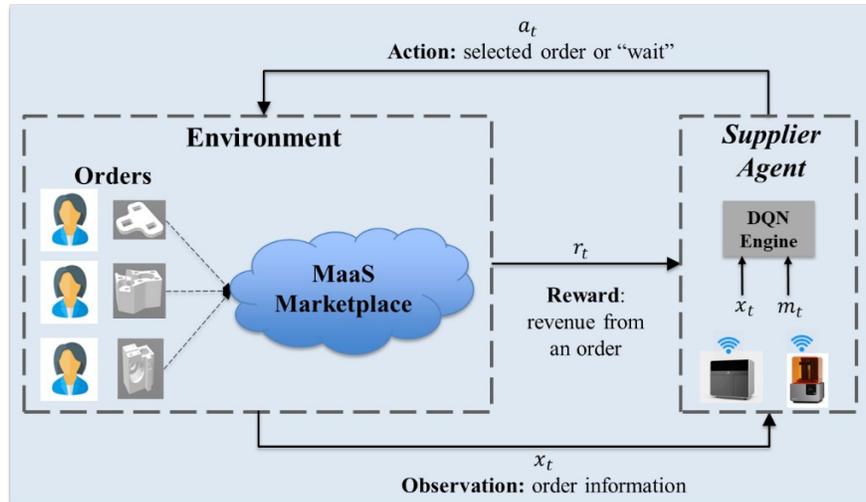

Figure 5-23: Agent and environment interactions in a MaaS marketplace

**State:** the state $s_t$ consists of attributes of available orders $x_t$ i.e. part volume, due date and required material, supplier attributes $m_t$ i.e. available capacities and material availability. The number of available orders in a period can vary resulting in a change in state dimension. Neural networks for function approximation require the state dimension to be fixed. Therefore, a fixed number of orders $\lambda_d$ are considered to be a part of the state and any additional orders are assigned to a queue. The length of the queue is considered as an additional parameter in the state definition.

**Action:** An action $a_t$ for the supplier agent is to choose an order to accept. As a supplier can accept multiple orders, the number of possible order combinations for a supplier can grow exponentially with the number of orders. Therefore, a supplier is allowed to choose a single order as an action to keep the number of actions linear in the number of orders. Here as well, the action space is non-stationary and to keep it fixed, the supplier is only allowed to choose from the set of orders considered in the state definition. Additionally, an action "wait" is added where the agent chooses to move to the next period and does not choose any of the available orders.

**Reward:** The reward function is designed as follows:

$$r_t = \begin{cases} p_i/h_i \text{ if a valid order is chosen} \\ -B \text{ if an invalid order is chosen} \\ -C * c_{t+1} \text{ if wait action is chosen} \end{cases}$$

where $p_i$ is the revenue generated from order $i$, $h_i$ is the production time of the order $i$, $c_{t+1}$ is the available capacity in the period $t+1$, $C$ is the penalty for each hour of capacity which goes waste (as the agent decided to move to the next period) and $B$ is the penalty for choosing an invalid order i.e. an order for which it does not have enough capacity to manufacture it by its due date. This reward function ensures that the supplier agent maximizes its revenue over a horizon $t=1T$ while utilizing the capacity to its fullest. It also discourages the supplier from choosing an invalid order.

Revenue $p_i$ from an order depends on its utility $u_i$ for the supplier. To determine $u_i$, expected utility theory [21, 22] is used as demonstrated in Chapter 4. $u_i$ depends on three factors – the urgency of required delivery, material availability for the order and the available capacity of the supplier. The closer the due date of the part, the lower the utility as the supplier has less time to fulfill the order. The higher the available capacity, the higher the utility and an order for which the required material is available has a higher utility. Revenue is considered to be utility dependent as follows:

$$p_i = bp_i + k * bp_i * (u\|i)^{-l}$$

$$bp_i = up_i * v_i$$

where $bp_i$ is the base price of the order which is determined by the volume of the part $v_i$ and price per unit volume $up_i$. $k$ and $l$ are constants which bound the increase in price with utility. The revenue and utility are assumed to have a nonlinear relationship with the supplier charging a higher price for a part with lower utility. Other price functions can also be considered. However, the price of the part ideally should be determined using historical data to capture the relationship between price and order/supplier attributes.

The deep Q-network (DQN) algorithm [5] is described in Figure 5-3. The agent observes the state and takes an action $a_t$ based on the $\varepsilon$-greedy policy. The policy chooses $a_t$ randomly with a probability $\varepsilon$, and based on the policy $\left(a_t = \underset{a}{argmax} \, Q(s_t|\theta)\right)$ with probability $1 - \varepsilon$. The random actions allow the agent to explore the environment whereas the actions based on the policy let the agent exploit the learned policy. The value of $\varepsilon$ starts at 1 and exponentially goes down to 0.01 over the learning duration. After the action is taken, the environment is updated and the reward $r_t$ and the next state $s_{t+1}$ is observed. The agent stores the transition $(s_t, a_t, r_t, s_{t+1})$ in the replay memory. To facilitate stable training of the DQN, we consider two networks – a predictor DQN and a target DQN. A set of transitions are sampled uniformly at random from replay memory. The predictor values $y_i$ are determined using predictor DQN and target values $\hat{y}_i$ are determined using target DQN considering a discount factor $\gamma$. The parameters $\theta$ of the predictor DQN are updated by minimizing the mean square loss $L(\theta)$ between the target and predictor values. Adam optimizer [23] is used for minimizing the loss function. After every $k$ iterations, the weights of the predictor

---

Initialize replay memory $D$

Use random weights $\theta$ to initialize the model

**for** $n=1$ to number of episodes do:

    Reset the environment and obtain the initial state $s_0$

    **for** every period $(t=1\,T$ **do**:

        the predictor DQN model observes the state $s_t$ and outputs action $a_t$ based on $\varepsilon$-greedy policy

        the simulator updates the environment and observes the reward $r_t$ and next state $s_{t+1}$

        the transition of agent $(s_t, a_t, r_t, s_{t+1})$ is stored in the replay memory $D$

        **for** $m = 1$ to $M$ **do:**

            sample $G\,U(D)$ transitions $(s_t, a_t, r_t, s_{t+1})$ from replay memory $D$

            **for** each transition in $G$ **do**:

                use the predictor DQN model to calculate $y = Q(s_t, a_t | \theta)$

                use the target DQN model to calculate $\hat{y} = r_t + \gamma \max_a Q(s_{t+1}, a | \theta')$

                minimize mean square loss $L(\theta) = E\left[(\hat{y} - y)^2\right]$ to update the parameters $\theta$ of

> the predictor DQN model
>
> **if** $\frac{n}{k}=0$ **do**:
>
> > copy parameters of predictor DQN network to target DQN network $(\theta=\theta')$

Figure 5-24: DQN algorithm for order acceptance in MaaS marketplace

---

Initialize states

**for** each period in an episode ($t=1\,T$ **do**:

> **Implement Policy:** supplier agents selects an action as per the $\varepsilon$-greedy policy. The action is either a valid order or an invalid order or "wait". The policy is implemented in a period until an invalid order or "wait" action is selected.
>
> **Update environment:**
>
> > **if** action is a valid order:
> >
> > > selected order is assigned and machine capacity is updated considering the production time of the order selected
> >
> > **if** action is invalid order or wait:
> >
> > > **Generate orders:** new orders are generated considering the distributions for order attributes such as material, due date and volume
> > >
> > > **Generate machine attributes:** capacity and material availability is generated for the next period
>
> **Determine Next State:** considering the available orders and machine attributes the next state for the supplier agent is generated

Figure 5-25: MaaS Marketplace Simulator

DQN are copied to the target DQN to ensure that the target values do not change frequently and the learning is stable.

The simulator used for testing the algorithm is defined in Figure 5-4. Orders arrive sequentially on the platform and in each period the platform takes an action to accept orders until it selects an invalid order or the action "wait". For each valid order selected, it updates the environment by updating its capacity and taking that order off from the available orders. An order from the queue moves to the set of available orders. Once, it selects an invalid order or "wait" action, the simulator moves to the next period. Waiting orders remain in the queue until a period before their due date (period in which they are due).

## 5.5 Simulation Environment

The proposed DQN method is compared against four baselines: Tabular Q-learning (TQ) where a table is used to store the $Q(s_t, a_t)$ values in contrast to the function approximation used in DQN. Since the states are stored in a table, the state definition has to be concise in order for the algorithm to be computationally tractable. State was considered to be a tuple of number of orders waiting to be assigned and the sum of available capacity rounded to the nearest integer. The second baseline is optimization using a rolling horizon approach (RHA) where the agent maximizes its revenue in each period given the constraints on capacity and due date. The third baseline is a greedy heuristic (GH) where, in each period, the agent selects a valid order which results in highest revenue per hour of capacity until there are no more valid orders available or it is out of capacity. The fourth baseline is a random algorithm which selects valid orders randomly in each period until there are no more valid orders available or it is out of capacity. The random seed in all algorithms is kept same to ensure a fair comparison.

The neural network in DQN consists of three fully connected hidden layers with 128, 64 and 32 neurons respectively. Each hidden layer uses a rectified linear unit as the activation function and the output layer uses a linear activation function. The algorithm is trained for 4,000 episodes with each episode consisting of 30 periods. Three orders arrive in each period and machine capacity arrives with a Poisson distribution having a rate of 8 hours per period (capped at 12 hours per period). Five different materials are considered (aluminum alloy, copper, titanium and two stainless steel alloys) and material availability is modelled as a binary variable with probability of availability 0.7, 0.4, 0.4, 0.5 and 0.3 respectively. Symmetric due date distribution provided in Table 5-2 is used. The constants for revenue function $k$ and $l$ are 0.5 each. The size of replay memory buffer is 20,000 and batch size is 500. The discount factor is 0.9 and the learning rate is 0.0001.

## 5.6 Results and Discussion

### 5.6.1 Comparison with baselines

The results shown in Table 5-1 demonstrate that DQN performs considerably better than the baselines. Note that DQN not only needs to learn the reward mapping with the state but also to utilize capacity efficiently over time. DQN also has a slightly higher order acceptance rate compared to the baselines. This can be attributed to the fact that the DQN, in order to maximize the revenue over time, chose smaller orders that provide higher revenue per hour leading to higher acceptance rate. It learns to ignore orders, which provide a lower per hour revenue and then reserve the available capacity for potentially better orders arriving in the future periods. TQ performs slightly better than RHA but significantly worse than DQN primarily because the state definition in TQ captures very limited environment information. Figure 5-5 represents the moving average (50 episodes) of the normalized revenue for DQN and the four baselines graphically demonstrating the gap between the performances of the algorithms.

Table 5-12: Mean and confidence intervals (CI) of normalized revenue and order acceptance rate of DQN vs baselines over 100 episodes

| Algorithm | Normalized Revenue (mean) | Normalized Revenue (CI) | Order acceptance rate (mean) | Order acceptance rate (CI) |
|---|---|---|---|---|
| **DQN** | 119.15 | 1.81 | 0.61 | 0.01 |
| **TQ** | 107.61 | 1.80 | 0.58 | 0.01 |
| **RHA** | 106.21 | 1.85 | 0.59 | 0.01 |
| **GH** | 101.95 | 1.69 | 0.54 | 0.01 |
| **Random** | 100.00 | 1.79 | 0.57 | 0.01 |

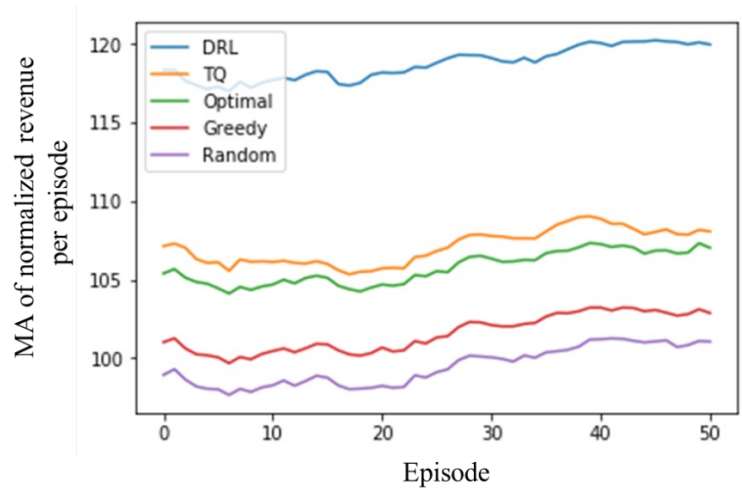

Figure 5-26: Moving average (50 episodes) of normalized revenue for 100 test episodes

### 5.6.2 Convergence behaviour of DQN and TQ

Figure 5-6 represents the moving average of normalized revenue per episode for the DQN algorithm as the learning progresses over a duration of 4,000 episodes. The algorithm starts with a performance worse than even the random allocation and improves as it learns the optimal policy. At around 1,000 iterations it surpasses all baselines and continues to improve until episode 2,000 where it starts to converge.

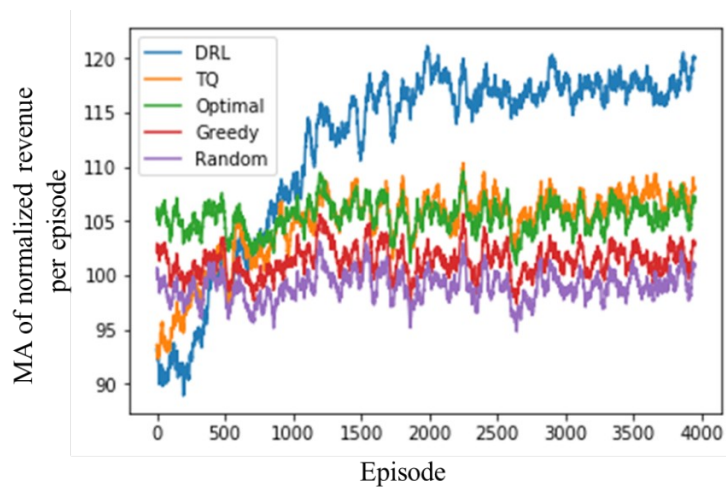

Figure 5-27: Convergence curve of DQN Model

Tabular Q (TQ) learning algorithm on the other hand requires a much higher number of iterations to converge. Figure 5-7 demonstrates the convergence curve for the size of the Q table i.e. the number of states encountered as the algorithm learns through trial and error. The number of states converge around 25,000 iterations but the $Q(s,a)$ values in the Q table need much larger number of episodes to converge. The algorithm does not converge even after running for 200,000 iterations as shown in Figure 5-8. As discussed in Section 5.5, TQ method suffers from two drawbacks – curse of dimensionality and the inability to generalize. Even after considering only two parameters (to avoid the curse of dimensionality) in the state definition, it took 25,000 episodes to reach almost all possible states at least once. Additionally, as this method cannot generalize, it has to visit each state enough number of times in order for the $Q(s,a)$ values to converge. Even with 200,000 iterations, the method could not accomplish that in contrast to DQN algorithm which required only 2,000 iterations because of its ability to generalize well.

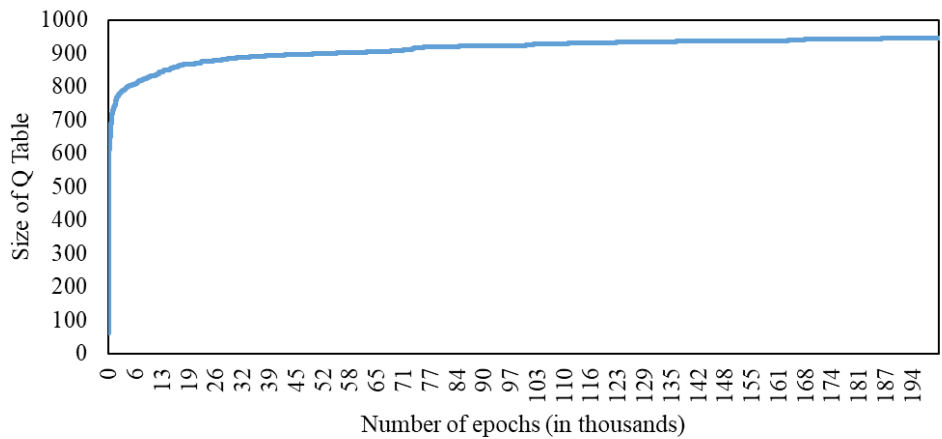

Figure 5-28: Convergence curve for size of the Q table

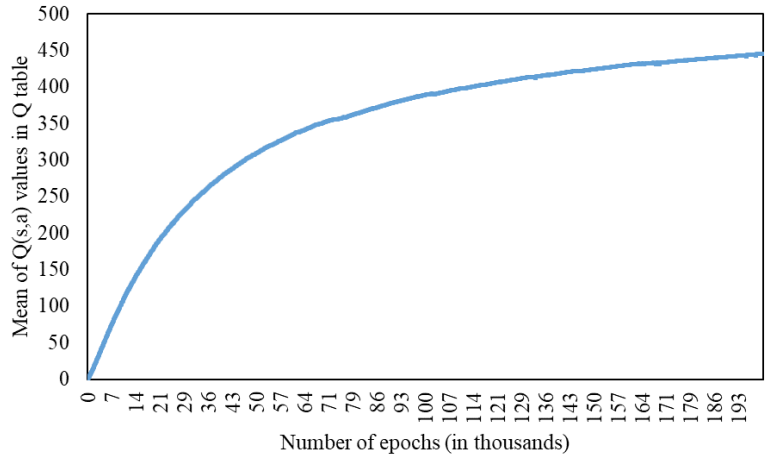

Figure 5-29: Convergence curve for mean values of Q table

### 5.6.3 Behavior of DQN under different distributions of machine capacity arrival

Since the system is operating under a capacity constrained environment, it is maximizing the revenue it can receive from the incoming orders. Unlike the myopic baselines (RHA, GA, RA), it rejects an order if the reward (revenue/ hour) from the order is less than expected reward it can gain from the future orders. Therefore, we evaluate the reward for the rejected orders under different capacity constrained environments. Figure 5-9 presents the average revenue/ hour for the rejected orders (orders for which the platform decided not to accept and wait for better orders in the future) for three different capacity arrival rates with a Poisson distribution. The results demonstrate that as the capacity gets constrained, the agent rejects the orders with higher revenue/ hour as expected. Under known distributions of incoming orders and capacity arrivals, a threshold reward can be determined where an agent can reject the orders below the threshold as proposed by [11]. However as the distributions change, the threshold would change accordingly as demonstrated in Figure 5-9. The DQN agent, however learns this strategy purely by trial and error to accept or reject orders as they come in.

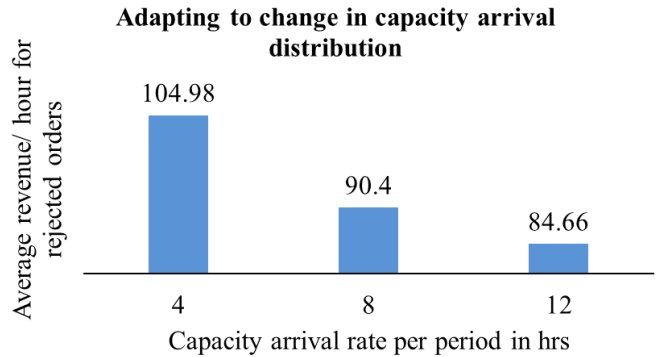

Figure 5-30: Average revenue/ hour of rejected orders for different capacity arrival rates

5.6.4 **Behavior of DQN under different due date distributions of arriving orders**

Similar to the behavior of DQN agent under different capacity distributions, we tested the average reward of rejected orders under different distributions of due dates of incoming orders. The distributions are presented in Table 5-2 and the results are presented in Figure 5-10. Under the left skewed distribution, majority of the orders have farther due dates and as we move from left skewed distribution to symmetric and then to right skewed, the incoming orders become urgent with closer due dates. As the urgency of incoming order increases, the competition for the available capacity increases and the agent learns to reject the orders with higher values compared to the case when most orders are less urgent and there is less competition for the available capacity.

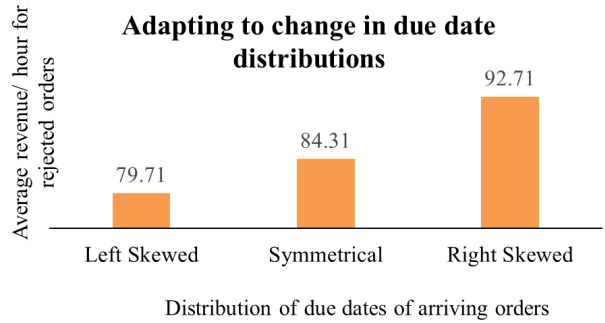

Figure 5-31: Average revenue/ hour of rejected orders for different due date distributions

Table 5-13: Distribution of due dates for orders arriving in period *t*

| Period | Probability of due date arrival | | |
|---|---|---|---|
| | **Left Skewed** | **Symmetric** | **Right Skewed** |
| $t+2$ | 0.05 | 0.1 | 0.4 |
| $t+3$ | 0.05 | 0.15 | 0.4 |
| $t+4$ | 0.05 | 0.25 | 0.05 |
| $t+5$ | 0.05 | 0.25 | 0.05 |
| $t+6$ | 0.4 | 0.15 | 0.05 |
| $t+7$ | 0.4 | 0.1 | 0.05 |

## 5.7 Conclusion

Advancements in cybermanufacturing technology has given rise to MaaS marketplaces where a large number of suppliers and customers participate in a network. These marketplaces operate in a highly dynamic and stochastic environment where orders and supplier capacities arrive over time. Moreover, conditions at suppliers' end change because of machine failures, order cancellations etc. In such an environment suppliers face this sequential decision making task of accepting or rejecting incoming orders from the platform. The suppliers need to take into consideration future arrival of orders and available capacities while making order acceptance decisions. This work proposes a method for online decision making where an agent learns to maximize value for a supplier in the platform over a period of time. The results demonstrate that the method learns purely from experience and performs significantly better than the baselines. Moreover, it is able to adapt to changing uncertainties in the marketplace. The method needs to be tested in a large environment with real datasets where the price quote for an order i.e. the value derived from an order and production time i.e. resources consumed, can both be estimated by historical data. Another key next step is to consider a multi agent system where both suppliers and customers participate as self-interested agents.

# Chapter 6 : Overall Summary, Research Contributions and Future Work

## 6.1 Overall Summary and Research Contributions

Digitization has transformed several industries such as ride sharing, hospitality and food delivery services but the manufacturing industry is yet to keep pace with the advancements in technology. Similar to online ride sharing and food delivery platforms, manufacturing-as-a-service marketplaces have emerged in the last decade. In order to participate on these platforms and remain competitive, the manufacturers need to embrace digitation and evolve too. Large manufacturers often have the tools and resources, which small and medium scale suppliers lack, to competitively participate on these platforms. Since, small and medium scale manufacturers contribute more than 70% of all US manufacturing, it is essential to develop tools to enable their participation in the digital manufacturing space.

The MaaS marketplaces operate in a centralized manner controlling key decision making where suppliers are boxed into a system of accepting orders at given prices. Decentralized operation of the marketplace brings in transparency leaving key decisions to the participants and supporting them with decision making tools. This dissertation has developed decision making tools which enable decentralization in decision making in MaaS marketplaces. This work is focused on two key decision making areas: pricing and matching.

In Chapter 2, this work proposes a machine learning approach to determine a competitive price for 3D printing services in an online 3D printing marketplace. The proposed method uses the features of a supplier such as customer reviews and supplier capabilities to predict a price range based on features and prices of other suppliers in the network. The analysis of data from

'3D Hubs' marketplace shows that price range of a supplier's listing can be successfully predicted using an array of features extracted from the supplier profiles. The support vector machine model built using the 3D Hubs dataset demonstrates an accuracy of 65% for suppliers in the US and 59% for suppliers in Europe. In a decentralized MaaS marketplace, suppliers determine the price charged to the designers and the platform takes a fee for each transaction on the marketplace. In such as business model, the developed pricing tool can be used by the suppliers to determine a competitive price range for the services they offer.

In Chapter 3, a reverse auction mechanism is proposed where designers bid with prices they are willing to pay for and the mechanism finds a qualified supplier which can match the stated bid price and order specifications. The novel two stage algorithm ensures that every supplier receives a chance of winning the game and there is no perceived direct competition among the suppliers (at least in the first round of supplier selection). This reverse auction type pricing model is a significant shift from existing service bureau marketplace practices. The approach also allows price of a particular service to be determined by the participants themselves – the client or the service provider and not by the middleware agent.

In Chapter 4, stable matching methods are introduced for resource allocation in a MaaS marketplaces and are empirically evaluated for different criteria of stability. The results demonstrate that enforcing stability introduces a slight loss in the social welfare of the platform i.e. total utility achieved by the platform. However, it ensures a sustainable marketplace in the long run. Among the proposed stable solutions, the approximately stable matching is recommended. It performs equally well in terms of utility, slightly worse in terms of stability but is significantly more time efficient compared to the maximum weight approximately stable matching.

In Chapter 5, a learning approach using deep Q-networks is considered for resource allocation in MaaS marketplaces. Conventional methods are myopic and do not adapt to changing uncertainties in the marketplace. In the proposed framework, a reinforcement learning agent learns to maximize revenue for a supplier in the platform over a period of time. The approach does considerably well compared to baselines such as a greedy heuristic and a pure optimization based approach. The agent learns the policy directly from experience and is able to adapt to different distributions of order arrival and machine capacity arrival in the marketplace.

## 6.2 Future Work

The work in this dissertation lays the foundation of a variety of frameworks for decision making in MaaS marketplaces. For pricing mechanisms (Chapter 2), the success of data mining-based models is only limited by availability of additional data from an online manufacturing marketplace. Historical sales data from a manufacturing marketplace could be used to determine the probability of winning an order at a specific price. In addition to supplier's features, order attributes such as part complexity (complex features in 3D Design), its due date, required quantity, required material, process and surface finish, and designer's attributes such as future potential of orders from the designer can also be considered for a price quote. Dynamic pricing models can be developed considering the impact of additional parameters such as demand forecast and raw material prices. Additional algorithms which are faster and can capture more complex relationships among predictor variables such as XG Boost [1], or deep neural networks can be developed.

In the reverse auction framework (Chapter 3), methods need to be developed to recommend threshold prices for the suppliers and bid recommendation for the designers. For stable matchings in Chapter 4, computationally efficient approaches such as a column generation method [2] to solve NP-hard integer programming formulations, particularly MWAS matching, need to be developed. Additional criteria for stability such as a blocking group consisting of multiple suppliers and designers need to be considered. In addition to weak stability, super and strong stability [3] can also be evaluated for participants as they have indifferent preferences. As the parts are printed in batches in 3D printing chambers, batching decisions [4] create additional complementarities between parts which lead to saving support material and eventually reduce printing costs for the supplier. These complementarities can also be considered in the three formulations proposed in Chapter 4. Introducing these complementarities will also require developing computationally efficient methods for pre-processing feasible combinations of contracts a supplier can accept. This is because the number of possible combinations increase exponentially with the number of feasible contracts. Finally, uncertainties in order arrivals and machine capacity arrivals need to be considered in order to improve the performance of the system in terms of posterior stability.

We considered price to be the only variable parameter in contracts between a supplier and designer. Additional parameters such as due dates and post processing methods can be considered to be varying among contracts as these parameters also enable negotiations between suppliers and designers. We used expected utility theory [5] to quantify the preferences of participants over contracts and rank them. A more practical approach to quantify the preferences of the participants over contracts is to use the historical transaction data to learn the preferences [6]. These marketplaces generate large amount of data which can be used to understand the participants' preferences and provide additional value to them.

We assumed that under stable matching participants will accept the matches performed by the platform. However, it is not necessary for the participants to accept the matches as they can be driven by factors not modelled in their preferences. Therefore, a framework where participants are presented with a menu of choices for acceptance can be developed.

The reinforcement learning method proposed in Chapter 5 needs to be tested with a larger number of suppliers. Each supplier having a DQN agent can be computationally expensive. Therefore, suppliers can be clustered based on their attributes and a single agent can represent all suppliers in a group for a distributed yet, computationally efficient decision making framework for suppliers. The method also needs to be tested with real datasets where the price quote for an order i.e. the value derived from an order and production time i.e. resources consumed, can both be estimated by historical data.

Finally, the methods on pricing and matching need to be tested in a combined framework [7] as price is one of the key considerations when the supplier makes a decision on order acceptance. The DRL agents can be modified so that the agent's action is to determine a price quote which would then impact the acceptance decision by the designers. Moreover, designers can also be considered active decision making agents for a truly multi agent MaaS marketplace enabling negotiations between agents.

Several other challenges faced by the MaaS marketplaces need to be resolved. Designing reputation systems [8] on which designers can rate services provided by the suppliers is key to sustain operations of the platform. Developing automated algorithms which can recommend technology such as 3D printing, injection molding or CNC machining to produce a design is required. 3D designs shared on the platform are often confidential and security of manufacturing data [9] is another dimension for research in these marketplaces. Automated validation of suppliers' machines [10] is necessary to maintain credibility of the participating suppliers on the platform.